\documentclass[journal]{IEEEtran}
\usepackage [latin1]{inputenc}%

\usepackage{amssymb}
\usepackage{cite}

\ifCLASSINFOpdf
\else
\fi
\usepackage{amsmath}
\usepackage{algorithmic}

\usepackage{array}
\usepackage{url}

\usepackage{amsthm}
\usepackage{booktabs}
\usepackage{graphicx}
\usepackage{epstopdf}

\usepackage{amssymb}
\usepackage{url}
\usepackage{amsfonts}
\usepackage{lineno}
\usepackage{subfigure}
\usepackage{multirow}
\usepackage{amsmath}
\usepackage{threeparttable}
\usepackage{algorithm}
\usepackage{algorithmic}
\usepackage{booktabs}
\usepackage{color}
\usepackage{multirow}
\usepackage{extarrows}
\usepackage{xcolor}
\usepackage{graphicx}
\usepackage{enumerate}

\usepackage[font=small,labelfont=bf,tableposition=top]{caption}
\usepackage{booktabs}
\usepackage{threeparttable}

\usepackage{float}

\newtheorem{theorem}{Theorem}
\newtheorem{prop}[theorem]{Proposition}

\renewcommand\tabcolsep{0.5pt}

\begin{document}
%
\title{Capped norm linear discriminant analysis and its applications}
%
%
%

\author{Jiakou~Liu, Xiong Xiong, Pei-Wei~Ren, Da~Zhao, Chun-Na~Li, Yuan-Hai~Shao
\IEEEcompsocitemizethanks{
\IEEEcompsocthanksitem Jiakou~Liu and Xiong Xiong are with the College of Management and Economics, Tianjin University, Tianjin, 300072, P.R.China. e-mails: liujiakou@163.com, xxpeter@tju.edu.cn.
\IEEEcompsocthanksitem Pei-Wei Ren is with the School of Science, Hainan University, Haikou, 570228, P.R.China. e-mail:  rpw19960717@163.com.
\IEEEcompsocthanksitem Da~Zhao, Chun-Na~Li and Yuan-Hai Shao are with the Management School, Hainan University, Haikou, 570228, P.R.China. e-mails: zhaoda@hainanu.edu.cn, na1013na@163.com, shaoyuanhai21@163.com. (Chun-Na~Li is the corresponding author)}
}


%
%

\markboth{Journal of \LaTeX\ Class Files,~Vol.~ , No.~ ,  ~ }%
{Shell \MakeLowercase{\textit{et al.}}: Capped norm linear discriminant analysis and its applications}
%



\maketitle

\begin{abstract}
Classical linear discriminant analysis (LDA) is based on squared Frobenious norm and hence is sensitive to outliers and noise. To improve the robustness of LDA, in this paper, we introduce capped $l_{2,1}$-norm of a matrix, which employs non-squared $l_2$-norm and ``capped" operation, and further propose a novel capped $l_{2,1}$-norm linear discriminant analysis, called CLDA. Due to the use of capped $l_{2,1}$-norm, CLDA can effectively remove extreme outliers and suppress the effect of noise data. In fact, CLDA can be also viewed as a weighted LDA. CLDA is solved through a series of generalized eigenvalue problems with theoretical convergency. The experimental results on an artificial data set, some UCI data sets and two image data sets demonstrate the effectiveness of CLDA.
\end{abstract}


%
\IEEEpeerreviewmaketitle

\section{Introduction}
The aim of dimensionality reduction is to embed high dimensional data into a low-dimensional space such that the most discriminative information is preserved. The projected data obtained by dimensionality reduction can be used in subsequent data mining tasks including classification, computer visualization, etc. In supervised learning, linear discriminant analysis (LDA) \cite{Fisher36, Fukunaga90} is one of the most useful and popular dimensionality reduction methods, and has been applied in many area, including multimodal dimensionality reduction \cite{ZhangZhaoChow12}, audiovisual speech recognition \cite{ZeilerNicheliMa16}, image recognition \cite{SongZhengWu10,RaghavendraAcharya16}, clustering \cite{LiShaoGuo19}, and tensor extension \cite{WangZhangLi17,YinMa20}.
LDA aims to learn a set of optimal projections to extract useful discriminative information, through maximizing the between-class distance and simultaneously minimizing the within-class distance in the projected space.

However, the construction of LDA was based on squared Frobenious norm (F-norm), or squared $l_2$-norm in essence. When facing data with noise or outliers, LDA will be the sensitive to them, which may in turn lead the drifting of projections away from the desired directions. To alleviate this problem, many researchers tried to improve the robustness of LDA.
For example, using subspace information \cite{ChenLiao00,YuYang01,SwetsWeng96,LaiMo18}, considering robust counterparts of sample means and covariance matrices \cite{Randles78,RLDA,HubertDriessen04, YuCaoJiang17}, using data uncertainty and optimizing for the worst-case \cite{KimMagnani05}, incorporating data local information \cite{LFDA,PLFDA,ZhangChow12,Okwonu}. As an effective robust replacement of squared $l_2$-norm, $l_1$-norm was usually used to resist outliers. By directly replacing $l_2$-norm with $l_1$-norm in LDA, several ratio form $l_1$-norm based LDA along with different solving algorithms and applications were studied \cite{ZhongZhang13,WangLu14,Liu17,YeYangLiu18}, for example, sum of $l_1$-norm based LDA \cite{ChenYangetal14}, constraint $l_1$-norm based LDA \cite{LiShao17,LiShao20,ZhangSunYe20}, and error minimized $l_1$-norm based LDA \cite{Zheng14,LiShaoWang19}.
Some $l_1$-based methods were also extended to $l_p$-norm ones \cite{OhKwak13,YeFuZhang18,LiShaoWang19-2} or corresponding matrix input LDAs \cite{LiShao15,ChenL12DLDA15,TrL12DLDA17,LiShang19}.

Since $l_1$-norm was not rotational invariant \cite{DingZhouHe06}, another robust and rotational invariant R$_1$-norm based LDA was studied in \cite{LiHuWang10}. R$_1$-norm of a matrix was defined as an $l_1$-norm sum of vector $l_2$-norms. R$_1$-norm used the $l_2$-norm rather than squared $l_2$-norm as basic norm, and considered the $l_1$-norm sum, which largely reduced the importance of outliers, and therefore had robustness.
R$_1$-norm was also known as $l_{2,1}$-norm in \cite{NieHuangCai10}, and was generalized to $l_{p,q}$-norm for arbitrary $p,q>0$. $l_{2,1}$-norm based LDA was also studied \cite{YangYeChen20}. However, though $l_{2,1}$-norm is robust, it still will suffer from the existence of odd points with extremely large norm. In this situation, even $l_{2,1}$-norm is not robust enough to reduce the effect of outliers. What if we can remove this influence or limit it within some boundary? In fact, this idea was employed in previous study to construct robust machine learning models, named capped norm. The central ideal of ``capped" operation is to add an upper bound on common norms. For example,
capped $l_2$-norm for robust feature learning \cite{LanHouYi16},
capped $l_{2,1}$-norm for regression and classification \cite{MaZhaoZhang18,ZhaoZhangZhan17}, capped $l_p$-norm for classification \cite{NieWangHuang17}, capped nuclear norm for matrix factorization or completion \cite{GaoNieCai15,ZhangYangChen18}, capped trace norm \cite{SunXiangYe13} for robust principal component analysis.
We found that though most of the above methods seemed to use different capped norms, and were applied to different machine learning models, most of them were essentially of the form
\begin{equation}\label{cap0}
\sum_{j=1}^s\min\{||\textbf{m}_j||_2,\epsilon\},
\end{equation}
where $\textbf{m}_j$, $j=1,2,\ldots,s$, was a series of column vectors and $\epsilon>0$.
From \eqref{cap0}, it can be seen that the capped operation can effectively remove outliers and noise beyond $\epsilon$ and benefit the input control.
Therefore, $\epsilon$ is a thresholding parameter for picking out extreme data outliers. If we do not cap the data with extremely large norm, it will affect the recognition result dramatically.

In \eqref{cap0}, if we deem $\textbf{m}_j$ as the $j$-th column of a matrix $\textbf{M}_{r\times s}$, $j=1,2,\ldots,s$, then we may define a ``norm" for the matrix $\textbf{M}_{r\times s}$. This motivates us to give the definition of capped norm of a matrix. As shown above, the existing works either define a capped norm directly as a loss or apply the capped operation on elements or vectors.
In fact, in \cite{SunHuangHuang20}, the authors defined a capped $l_{2,1}$-norm of a matrix. In this paper, we introduce a general capped norm of a matrix, named capped $l_{p,q}$-norm.
In specific, given some $\epsilon>0$ and $p,q>0$, we formally define a capped $l_{p,q}$-norm of a matrix $\textbf{M}_{r\times s}$ as
\begin{equation}\label{cappedlpq}
\|\textbf{M}\|_{cap\,{p,q}}=\left(\sum_{j=1}^s\min\{||\textbf{m}_j||_p^q,\epsilon\}\right)^{\frac{1}{q}},
\end{equation}
where $p,q>0$.
It should be noted that even we call them norms, capped $l_{2,1}$-norm and capped $l_{p,q}$-norm are not actual norms. In fact, as $T\ell_1$-norm \cite{YangShaoLi20}, they do not satisfy homogeneity. However, when $p,q\geq 1$, it satisfies the positive scalability and triangle inequality. Clearly, $\|\textbf{M}\|_{cap\,{p,q}}\geq 0$
and it can be easily verified that
\begin{equation}\label{cappedlpq}
\|\textbf{A+B}\|_{cap\,{p,q}}\leq \|\textbf{A}\|_{cap\,{p,q}}+\|\textbf{B}\|_{cap\,{p,q}}.
\end{equation}
However, as $Tl_1$-norm, it does not affect the robustness property, and we still call it a norm.

\begin{figure}[htbp]
\centering
\includegraphics[width=0.4\textwidth]{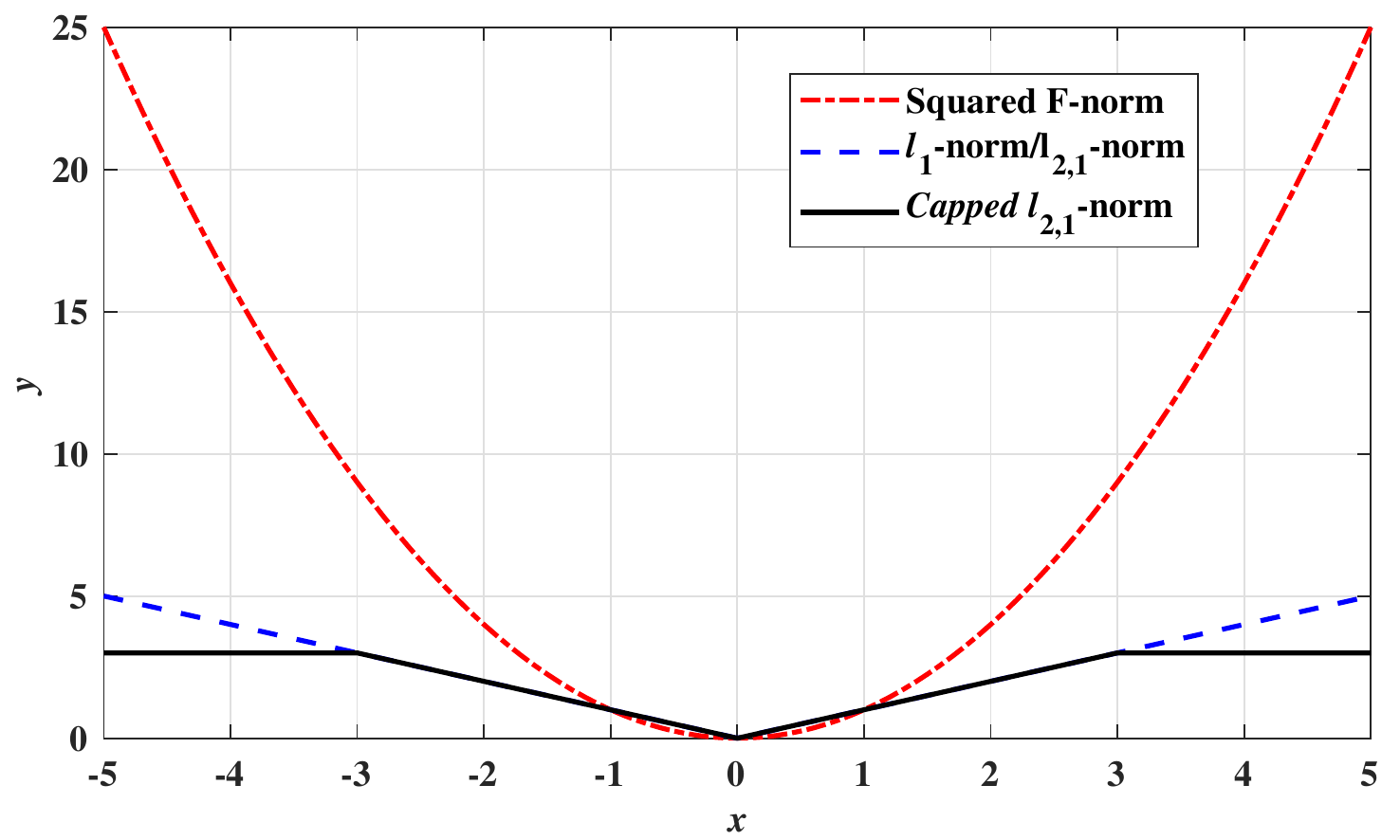}\hfill
\caption{Illustration of capped $l_{2,1}$-norm.}
\label{Fignorm}
\end{figure}

Considering the robustness of vector $l_2$-norm and $l_1$-norm, in this paper we only consider matrix capped $l_{2,1}$-norm, whose definition is based on vector $l_2$-norm and $l_1$-norm.
Compared to Frobenius norm (F-norm) which is defined as $\|\textbf{M}\|_F=\sqrt{\sum_{j=1}^s\sum_{i=1}^rm_{ij}^2}=\sqrt{\sum_{j=1}^s||\textbf{m}_j||_2^2}$, squared F-norm, and $l_{2,1}$-norm which is defined as $\|\textbf{M}\|_{2,1}=\sum_{j=1}^s\sqrt{\sum_{i=1}^r m_{ij}^2}=\sum_{j=1}^s||\textbf{m}_j||_2$,
capped $l_{2,1}$-norm can effectively suppress the effect of outliers and resist noise. The illustration of capped $l_{2,1}$-norm is demonstrated in Fig.\ref{Fignorm}.
In reality, while data outliers exist, feature noise can also exist.
Suppose $\textbf{M}$ is a data matrix with each column being a data sample. Since capped $l_{2,1}$-norm is summed over all ``capped" data samples, it is robust to outliers. In addition, since capped $l_{2,1}$-norm adopts $l_2$-norm as a basic norm that acts on the $j$-th data sample $\textbf{m}_j$, it is also robust to feature noise.

In this paper, based on the above defined capped $l_{2,1}$-norm, we propose a capped $l_{2,1}$-norm linear discriminant analysis (CLDA). Compared with classical LDA which utilized squared F-norm, the proposed CLDA is robust to outliers and noise. In specific, CLDA has following characteristics:

(i) A capped $l_{p,q}$-norm for arbitrary $p,q>0$ of a matrix is formally introduced, and a novel linear discriminant analysis based on capped $l_{2,1}$-norm named CLDA is proposed.

(ii) The capped $l_{2,1}$-norm used in CLDA makes it robust to data outliers and feature noise.

(iii) An effective algorithm is designed to solve the proposed non-smooth and non-convex optimization problem. The theoretical analysis of the proposed algorithm is also given.

(iv) Experimental results on an artificial data, some UCI data and two image data demonstrate the effectiveness of CLDA.

The paper is organized as follows. Section \ref{secRelated} briefly reviews LDA.
Section \ref{secCLDA} proposes CLDA and its corresponding analysis. Section \ref{secExp} makes comparisons of CLDA with its related approaches. Concluding remarks are given in Section \ref{secCon}.

The notation of the paper is listed as follows. All vectors are column ones, and vectors and matrices are shown in bold. We consider the training data set $T=\{\textbf{x}_1,\,\textbf{x}_2,\ldots,\,\textbf{x}_N\}$ with associated class labels $y_1,\,y_2,\ldots,\,y_N$ belonging to $\{1,\,2,\,\ldots,c\}$, where $\textbf{x}_l\in\mathbb{R}^n$ for $l=1,\,2,\,\ldots,N$.
Write $\textbf{X}=(\textbf{x}_1,\,\textbf{x}_2,\ldots,\,\textbf{x}_N)\in\mathbb{R}^{n\times N}$ as the corresponding data matrix of $T$.
Assume the $i$-th class contains $N_{i}$ samples, $i=1,2,\ldots,c$. Then $\sum\limits_{i=1}^{c}N_i=N$. Let $\overline{\textbf{x}}=\frac{1}{N}\sum\limits_{l=1}^{N}\textbf{x}_l$ be the mean of all samples and ${\overline{\textbf{x}}}_i=\frac{1}{N_i}\sum\limits_{j=1}^{N_i}\textbf{x}_{i}^j$ be the mean of samples in the $i$-th class, where $\textbf{x}_{i}^j$ is the $j$-th sample in the $i$-th class, $i=1,2,\ldots,c$, $j=1,2,\ldots,N_i$.
For simplicity, we write $\|\cdot\|_2$ as $\|\cdot\|$.

\section{Linear discriminant analysis}\label{secRelated}

LDA performs dimensionality reduction by seeking a linear transformation matrix
$\textbf{W}\in\mathbb{R}^{n\times d}$, $d\leq n$ such that the between-class scatter distance is maximized and meanwhile the within-class scatter distance is minimized in the projected space. In specific, LDA formulates as
\begin{equation}\label{LDA0}
\begin{split}
\underset{\textbf{W}}{\max}~~&\frac{\sum\limits_{i=1}^{c}N_i\|\textbf{W}^T(\overline{\textbf{x}}_{i}-\overline{\textbf{x}})\|^2}{\sum\limits_{i=1}^{c}\sum\limits_{j=1}^{N_i}\|\textbf{W}^T(\textbf{x}_{i}^j-\overline{\textbf{x}}_{i})\|^2}.
\\
\end{split}
\end{equation}
Write
\begin{equation}\label{Xb}
\textbf{H}_b=(\sqrt{N_1}(\overline{\textbf{x}}_{1}-\overline{\textbf{x}}),\sqrt{N_2}(\overline{\textbf{x}}_{2}-\overline{\textbf{x}}),\cdots,\sqrt{N_c}(\overline{\textbf{x}}_{c}-\overline{\textbf{x}}))
\in\mathbb{R}^{n\times c}
\end{equation}
and
\begin{equation}\label{Xw}
\textbf{H}_w=((\textbf{x}_{1}^1-\overline{\textbf{x}}_{1}),\cdots,(\textbf{x}_{1}^{N_1}-\overline{\textbf{x}}_{1}),\cdots,(\textbf{x}_{c}^1-\overline{\textbf{x}}_{c}),\cdots,(\textbf{x}_{c}^{N_c}-\overline{\textbf{x}}_{c}))
\in\mathbb{R}^{n\times N},
\end{equation}
then \eqref{LDA0} is equivalent to
\begin{equation}\label{LDA}
\underset{\textbf{W}}{\max}~~
\frac{\|\textbf{W}^T\textbf{H}_b\|_F^2}{\|\textbf{W}^T\textbf{H}_w\|_F^2}.
\end{equation}

The optimal solution $\textbf{W}=(\textbf{w}_1,\ldots,\textbf{w}_d)\in\mathbb{R}^{n\times d}$ of \eqref{LDA0} can be obtained from a generalized problem $\textbf{S}_b\textbf{w}=\lambda \textbf{S}_w\textbf{w}$ with $\lambda\not=0$, where $\textbf{S}_b$ and $\textbf{S}_w$ are between-class scatter matrix and within-class scatter matrix defined by
\begin{equation}\label{SB}
\textbf{S}_b=\frac{1}{N}\textbf{H}_b\textbf{H}_b^T
\end{equation}
and
\begin{equation}\label{SW} \textbf{S}_w=\frac{1}{N}\textbf{H}_w\textbf{H}_w^T
\end{equation}
respectively.
If $\textbf{S}_w$ is nonsingular, then $\textbf{W}$ is given by the first $d$ largest eigenvalues of $(\textbf{S}_w)^{-1}\textbf{S}_b$.

\section{The proposed capped $l_{2,1}$-norm linear discriminant analysis}\label{secCLDA}

\subsection{Problem formulation}
As we see, in LDA, the within-class distance and between-class distance are based on squared F-norm, which makes LDA sensitive to outliers and noise. The squared operation on $F$-norm or $l_2$-norm metric will enlarge the effect of outliers and noise. If we discard the square operation and consider just F-norm or $l_2$-norm, the robustness will be improved \cite{LiLiGao17}. However, even F-norm or $l_2$-norm may lose control to some outliers with large norms. In this situation, it is necessary to set a bar to remove outliers. This motivates us to introduce the aforementioned capped $l_{2,1}$-norm into LDA and construct a novel capped $l_{2,1}$-norm based LDA (CLDA).
In specific, the proposed CLDA has the following optimization problem
\begin{equation}\label{CLDA00}
\underset{\textbf{W}}{\min}~~
\frac{\|\textbf{W}^T\textbf{H}_w\|_{cap\,{2,1}}}{\|\textbf{W}^T\textbf{H}_b\|_{cap\,{2,1}}}
=\frac{\sum\limits_{i=1}^{c}\sum\limits_{j=1}^{N_i}\min\left(\|\textbf{W}^T(\textbf{x}_{i}^j-\overline{\textbf{x}}_{i})\|,\epsilon\right)}
{\sum\limits_{i=1}^{c}\min\left(\|\sqrt{N_i}\textbf{W}^T(\overline{\textbf{x}}_{i}-\overline{\textbf{x}})\|,\epsilon\right)},
\end{equation}
where $\epsilon>0$ is a thresholding parameter.
It can be seen that both the within-class distance and between-class distance are measured by capped $l_{2,1}$-norm. In these two capped $l_{2,1}$-norm terms, the application of $l_2$-norm on projected data $\textbf{W}^T(\textbf{x}_{i}^j-\overline{\textbf{x}}_{i})$ and $\sqrt{N_i}\textbf{W}^T(\overline{\textbf{x}}_{i}-\overline{\textbf{x}})$ reduces the negative influence of noise and outliers, and the $l_1$-norm sum over the ``capped" projected data restricted by parameter $\epsilon$ removes the effect of extreme data outliers.
To solve \eqref{CLDA00}, we first recast it to an equivalent form
\begin{equation}\label{CLDA}
\begin{split}
\underset{\textbf{W}}{\min}&~~
{\sum\limits_{i=1}^{c}\sum\limits_{j=1}^{N_i}\min(\|\textbf{W}^T(\textbf{x}_{i}^j-\overline{\textbf{x}}_{i})\|,\epsilon)}
\\
\hbox{s.t.\ }& {\sum\limits_{i=1}^{c}\min(\|\sqrt{N_i}\textbf{W}^T(\overline{\textbf{x}}_{i}-\overline{\textbf{x}})\|,\epsilon)}=1.
\end{split}
\end{equation}

The above formulation of CLDA in fact can be viewed as a formally generalized eigenvalue optimization problem.
Define
$F_{ij}=\|\textbf{W}^T(\textbf{x}_{i}^j-\overline{\textbf{x}}_{i})\|^{-1}
\cdot Ind(\|\textbf{W}^T(\textbf{x}_{i}^j-\overline{\textbf{x}}_{i})\|\leq\epsilon)$, where $Ind(\cdot)$ is the indicator function satisfying $Ind(\|\textbf{W}^T(\textbf{x}_{i}^j-\overline{\textbf{x}}_{i})\|\leq\epsilon)=1$ if $\|\textbf{W}^T(\textbf{x}_{i}^j-\overline{\textbf{x}}_{i})\|\leq \epsilon$, and 0 otherwise,
and define $\textbf{F}\in\mathbb{R}^{N\times N}$ as the diagonal matrix with its diagonal element $F_{ij}$, $i=1,2,\ldots,c,~,j=1,2,\ldots,N_i$.
Then
\begin{equation}\label{CLDAobj}
\begin{split}
&\underset{\textbf{W}}{\min}~~\sum\limits_{i=1}^{c}\sum\limits_{j=1}^{N_i}\min(\|\textbf{W}^T(\textbf{x}_{i}^j-\overline{\textbf{x}}_{i})\|,\epsilon)\\
=~&\underset{\textbf{W}}{\min}~~\sum\limits_{i=1}^{c}\sum\limits_{j=1}^{N_i} F_{ij}\|\textbf{W}^T(\textbf{x}_{i}^j-\overline{\textbf{x}}_{i})\|^2\\
=~&\underset{\textbf{W}}{\min}~~\hbox{tr}(\textbf{W}^T\textbf{H}_w\textbf{F}\textbf{H}_w^T\textbf{W})\\
=~&\underset{\textbf{W}}{\min}~~\hbox{tr}(\textbf{W}^T\textbf{S}_1\textbf{W}),
\end{split}
\end{equation}
where $\textbf{S}_1=\textbf{H}_w\textbf{F}\textbf{H}_w^T$.

Similarly, denote $G_{i}=\|\sqrt{N_i}\textbf{W}^T(\overline{\textbf{x}}_{i}-\overline{\textbf{x}})\|^{-1}
\cdot Ind(\sqrt{N_i}\|\textbf{W}^T(\overline{\textbf{x}}_{i}-\overline{\textbf{x}})\|\leq\epsilon)$ and
$\textbf{G}\in\mathbb{R}^{c\times c}$ as the diagonal matrix with its $(i,i)$-th element $G_{i}$. Then the left side of constraint of \eqref{CLDA} is equivalent to
\begin{equation}\label{CLDAcons}
\sum\limits_{i=1}^{c} G_{i}\|\sqrt{N_i}\textbf{W}^T(\overline{\textbf{x}}_{i}-\overline{\textbf{x}})\|^2=\hbox{tr}(\textbf{W}^T\textbf{S}_2\textbf{W}),
\end{equation}
where $\textbf{S}_2=\textbf{H}_b\textbf{G}\textbf{H}_b^T$.
Therefore, CLDA \eqref{CLDA} can be recast as
\begin{equation}\label{CLDA1}
\begin{split}
\underset{\textbf{W}}{\min}~&\hbox{tr}(\textbf{W}^T\textbf{S}_1\textbf{W})\\
\hbox{s.t.~~}\hbox{tr}&(\textbf{W}^T\textbf{S}_2\textbf{W})=\Delta,
\end{split}
\end{equation}
where $\Delta$ is some constant.
By observing \eqref{CLDA1}, we may deem CLDA as a generalized eigenvalue optimization problem in form, since $\textbf{S}_1$ and $\textbf{S}_2$ are both related to $\textbf{W}$. We call $\textbf{S}_1$ as within-class scatter and $\textbf{S}_2$ as between-class scatter of CLDA.

It should be noted that the within-class scatter $\textbf{S}_1$ and between-class scatter $\textbf{S}_2$ in CLDA are in fact weighted $\textbf{S}_w$ and $\textbf{S}_b$ in LDA, and $\textbf{F}$ and $\textbf{G}$ are their corresponding weighting matrices. These weights make CLDA robust to outliers and noise.
On one hand, non-squared $l_2$-norm brings robustness to feature noise. One the other hand, for projected data that have norm larger than $\epsilon$ will be ignored since the corresponding weight is 0, and these data are considered as outliers. For projected data having large norms but less than $\epsilon$, small weights are given to these data, and these data may be noise data. Therefore, CLDA is robust to outliers and noise.

\subsection{The solving algorithm of CLDA and convergence analysis}

In the following, we solve \eqref{CLDA1}. For fixed $\textbf{S}_1$ and $\textbf{S}_2$, \eqref{CLDA1} is a generalized eigenvalue problem.
Clearly, $\textbf{S}_1$ and $\textbf{S}_2$ depend on $\textbf{F}$ and $\textbf{G}$ respectively, and hence depend on $\textbf{W}$.
Therefore, to solve this problem, we employ an iterative technique.
In specific, we first initialize $\textbf{W}^{(0)}$ as the first $d$ columns of the identity $\textbf{I}\in\mathbb{R}^{n\times n}$. Then in the $t$-th iteration,
$\textbf{S}_1^{(t)}$ and $\textbf{S}_2^{(t)}$ are computed according to
\begin{equation}\label{CLDAS1S2}
\begin{split}
\textbf{S}_1^{(t)}&=\textbf{H}_w\textbf{F}^{(t)}\textbf{H}_w^T\\
\textbf{S}_2^{(t)}&=\textbf{H}_b\textbf{G}^{(t)}\textbf{H}_b^T,
\end{split}
\end{equation}
where the diagonal elements of $\textbf{F}^{(t)}$ and $\textbf{G}^{(t)}$ are given by $$F_{ij}^{(t)}=\|(\textbf{W}^{(t)})^T(\textbf{x}_{i}^j-\overline{\textbf{x}}_{i})\|^{-1}
\cdot Ind(\|(\textbf{W}^{(t)})^T(\textbf{x}_{i}^j-\overline{\textbf{x}}_{i})\|\leq\epsilon)$$ and $$G_{i}^{(t)}=\|\sqrt{N_i}(\textbf{W}^{(t)})^T(\overline{\textbf{x}}_{i}-\overline{\textbf{x}})\|^{-1}
\cdot Ind(\|\sqrt{N_i}(\textbf{W}^{(t)})^T(\overline{\textbf{x}}_{i}-\overline{\textbf{x}})\|\leq\epsilon).$$

After obtaining $\textbf{S}_1^{(t)}$ and $\textbf{S}_2^{(t)}$, the optimal solution $\textbf{W}^{(t+1)}$ of \eqref{CLDA1} is computed by solving the following problem
\begin{equation}\label{CLDAW}
\begin{split}
\textbf{W}^{(t+1)}&=\underset{\textbf{W}}{\arg\min}~~\hbox{tr}(\textbf{W}^T\textbf{S}_1^{(t)}\textbf{W})\\
&\hbox{s.t.\ }{\hbox{tr}(\textbf{W}^T\textbf{S}_2^{(t)}\textbf{W})}=\Delta.
\end{split}
\end{equation}
The optimal solution $\textbf{W}^{(t+1)}$ of \eqref{CLDAW} can be given by the first $d$ eigenvectors corresponding to first $d$ smallest nonzero eigenvalues of the generalized eigenvalue problem
$\textbf{S}_1^{(t)}\textbf{w}=\lambda\textbf{S}_2^{(t)}\textbf{w}$, where $\lambda$ is its eigenvalue.
After reaching maximum iteration number or convergence, the optimal $\textbf{W}^*$ is set to $\textbf{W}^{(t+1)}$. After obtaining $\textbf{W}^*$, for a new coming sample $\textbf{x}\in\mathbb{R}^{n}$, its representation in low dimension space is $\widehat{\textbf{x}}=(\textbf{W}^*)^T\textbf{x}$.

The solving procedure is summarized in Algorithm 1.
\begin{center}
\begin{tabular}{lllll}
\toprule
\noindent{$\mathbf{Algorithm~1.}$~ CLDA solving Algorithm}\\
\midrule
\textbf{Input}: Training data set $T=\{\textbf{x}_1,\,\textbf{x}_2,\ldots,\,\textbf{x}_N\}$, reduced\\
dimension $d\leq n$,
maximum iteration number $Itmax$, and\\
parameter $\epsilon>0$.\\
\textbf{Process}: \\
\quad{1. }Initialize $\textbf{W}^{(0)}$ as the first $d$ columns of the identity\\
matrix $\textbf{I}\in\mathbb{R}^{n\times n}$;\\
\quad{2. }\textbf{For $t=0,1,2,\ldots$, repeat}\\
\quad\quad\quad Compute $\textbf{S}_1^{(t)}$ and $\textbf{S}_2^{(t)}$ by \eqref{CLDAS1S2};\\
\quad\quad\quad Compute $\textbf{W}^{(t+1)}$ by solving \eqref{CLDAW};\\
\quad\quad\quad Set $t=t+1$;\\
\quad\quad\textbf{Until} reaching maximum iteration number $Itmax$ or\\ convergence.\\
\textbf{Output}: $\textbf{W}^*=\textbf{W}^{(t+1)}$ and $\widehat{\textbf{x}}=(\textbf{W}^*)^T\textbf{x}$.\\
\bottomrule
\end{tabular} \label{Algorithm1}
\end{center}
We have the following convergency result about Algorithm 1.
\begin{prop}
The procedures of CLDA shown in Algorithm 1 monotonically decrease the objective of \eqref{CLDA} in each step and converge to a local optimum.
\end{prop}
{\bf Proof:}
Let $K^{(t)}$ be the index set of $i=1,2,\ldots,c,~j=1,2,\ldots,N_i$ that satisfies $\|(\textbf{W}^{(t)})^T(\textbf{x}_{i}^j-\overline{\textbf{x}}_{i})\|\leq\epsilon$,
that is $K^{(t)}=\{(i,j):\|(\textbf{W}^{(t)})^T(\textbf{x}_{i}^j-\overline{\textbf{x}}_{i})\|\leq\epsilon\}$,
$t=0,1,2,\ldots$.
Denote $|K^{(t)}|$ as the number of elements in $K^{(t)}$.
Since $\textbf{W}^{(t+1)}$ minimizes \eqref{CLDAW},
\begin{equation}\label{FfixJineq2}
\begin{split}
&\sum\limits_{(i,j)\in K^{(t)}} F_{ij}^{(t)}\|(\textbf{W}^{(t+1)})^T(\textbf{x}_{i}^j-\overline{\textbf{x}}_{i})\|^2\\
\leq&\sum\limits_{(i,j)\in K^{(t)}} F_{ij}^{(t)}\|(\textbf{W}^{(t)})^T(\textbf{x}_{i}^j-\overline{\textbf{x}}_{i})\|^2.
\end{split}
\end{equation}
On one hand, note that when $(i,j)\in K^{(t)}$, $F_{ij}^{(t)}=\|\textbf{W}^T(\textbf{x}_{i}^j-\overline{\textbf{x}}_{i})\|^{-1}$, we have
\begin{equation}\label{FfixJineq3}
\begin{split}
&\sum\limits_{(i,j)\in K^{(t)}} \frac{\|(\textbf{W}^{(t+1)})^T(\textbf{x}_{i}^j-\overline{\textbf{x}}_{i})\|^2}
{2\|(\textbf{W}^{(t)})^T(\textbf{x}_{i}^j-\overline{\textbf{x}}_{i})\|}\\
\leq&\sum\limits_{(i,j)\in K^{(t)}} \frac{\|(\textbf{W}^{(t)})^T(\textbf{x}_{i}^j-\overline{\textbf{x}}_{i})\|^2}
{2\|(\textbf{W}^{(t)})^T(\textbf{x}_{i}^j-\overline{\textbf{x}}_{i})\|}.
\end{split}
\end{equation}
From the inequality $\sqrt{a}-\frac{a}{2\sqrt{b}}\leq \sqrt{b}-\frac{b}{2\sqrt{b}}$ for arbitrary $a,b>0$, we have
\begin{equation}\label{FfixJineq4}
\begin{split}
&\sum\limits_{(i,j)\in K^{(t)}}\left( {\|(\textbf{W}^{(t+1)})^T(\textbf{x}_{i}^j-\overline{\textbf{x}}_{i})\|}-
\frac{\|(\textbf{W}^{(t+1)})^T(\textbf{x}_{i}^j-\overline{\textbf{x}}_{i})\|^2}
{2\|(\textbf{W}^{(t)})^T(\textbf{x}_{i}^j-\overline{\textbf{x}}_{i})\|}\right)\\
\leq &\sum\limits_{(i,j)\in K^{(t)}}\left(
{\|(\textbf{W}^{(t)})^T(\textbf{x}_{i}^j-\overline{\textbf{x}}_{i})\|}-
\frac{\|(\textbf{W}^{(t)})^T(\textbf{x}_{i}^j-\overline{\textbf{x}}_{i})\|^2}
{2\|(\textbf{W}^{(t)})^T(\textbf{x}_{i}^j-\overline{\textbf{x}}_{i})\|}\right)\\
\end{split}
\end{equation}
By adding \eqref{FfixJineq3} and \eqref{FfixJineq4} together, we obtain
\begin{equation}\label{FfixJineq5}
\begin{split}
\sum\limits_{(i,j)\in K^{(t)}} {\|(\textbf{W}^{(t+1)})^T(\textbf{x}_{i}^j-\overline{\textbf{x}}_{i})\|}
\leq \sum\limits_{(i,j)\in K^{(t)}}
{\|(\textbf{W}^{(t)})^T(\textbf{x}_{i}^j-\overline{\textbf{x}}_{i})\|}.
\end{split}
\end{equation}
Now adding term $(N-|K^{(t)}|)\epsilon$ to both sides of the above inequality, it follows
\begin{equation}\label{FfixJineq6}
\begin{split}
&\sum\limits_{(i,j)\in K^{(t)}} {\|(\textbf{W}^{(t+1)})^T(\textbf{x}_{i}^j-\overline{\textbf{x}}_{i})\|}+(N-|K^{(t)}|)\epsilon\\
\leq&\sum\limits_{(i,j)\in K^{(t)}}
{\|(\textbf{W}^{(t)})^T(\textbf{x}_{i}^j-\overline{\textbf{x}}_{i})\|}+(N-K^{(t)}|)\epsilon.
\end{split}
\end{equation}
On the other hand, since
$K^{(t+1)}=\{(i,j):\|(\textbf{W}^{(t+1)})^T(\textbf{x}_{i}^j-\overline{\textbf{x}}_{i})\|\leq\epsilon\}$,
it can be verified that
\begin{equation}\label{FfixJineq7}
\begin{split}
&\sum\limits_{(i,j)\in K^{(t+1)}} {\|(\textbf{W}^{(t+1)})^T(\textbf{x}_{i}^j-\overline{\textbf{x}}_{i})\|}+(N-|K^{(t+1)}|)\epsilon\\
\leq& \sum\limits_{(i,j)\in K^{(r)}}
{\|(\textbf{W}^{(t+1)})^T(\textbf{x}_{i}^j-\overline{\textbf{x}}_{i})\|}+(N-|K^{(r)}|)\epsilon
\end{split}
\end{equation}
for arbitrary $r=1,2,\ldots,t$. In fact, if $K^{(r)}=\emptyset$, \eqref{FfixJineq7} is obvious true from the definition of $K^{(t+1)}$.
If $K^{(r)}\not=\emptyset$, we first write $K^{(r)}=K^{(r')}\bigcup(K^{(r)}\backslash K^{(r')})$, where $K^{(r')}\subseteq K^{(t+1)}$. Then for $(i,j)\in K^{(r)}\backslash K^{(r')}$, $\|(\textbf{W}^{(t+1)})^T(\textbf{x}_{i}^j-\overline{\textbf{x}}_{i})\|\geq \epsilon$.
Therefore,
\begin{equation}\label{FfixJineq8}
\begin{split}
&\sum\limits_{(i,j)\in K^{(r)}}
{\|(\textbf{W}^{(t+1)})^T(\textbf{x}_{i}^j-\overline{\textbf{x}}_{i})\|}+(N-|K^{(r)}|)\epsilon\\
=&\sum\limits_{(i,j)\in K^{(r')}}
{\|(\textbf{W}^{(t+1)})^T(\textbf{x}_{i}^j-\overline{\textbf{x}}_{i})\|}\\
&+\sum\limits_{(i,j)\in K^{(r)}\backslash K^{(r')}}
{\|(\textbf{W}^{(t+1)})^T(\textbf{x}_{i}^j-\overline{\textbf{x}}_{i})\|}+(N-|K^{(r)}|)\epsilon\\
\geq&\sum\limits_{(i,j)\in K^{(r')}}
{\|(\textbf{W}^{(t+1)})^T(\textbf{x}_{i}^j-\overline{\textbf{x}}_{i})\|}
+(|K^{(r)}|-|K^{(r)}\cap K^{(r')}|)\epsilon\\
&+(N-|K^{(r)}|)\epsilon\\
\geq&\sum\limits_{(i,j)\in K^{(r')}}
{\|(\textbf{W}^{(t+1)})^T(\textbf{x}_{i}^j-\overline{\textbf{x}}_{i})\|}
+(N-|K^{(r')}|)\epsilon\\
\geq&\sum\limits_{(i,j)\in K^{(t+1)}}
{\|(\textbf{W}^{(t+1)})^T(\textbf{x}_{i}^j-\overline{\textbf{x}}_{i})\|}
+(N-|K^{(t+1)}|)\epsilon.
\end{split}
\end{equation}
The last inequality of \eqref{FfixJineq8} follows from the fact that
\begin{equation}\label{FfixJineq9}
\begin{split}
&\sum\limits_{(i,j)\in K^{(t+1)}}
{\|(\textbf{W}^{(t+1)})^T(\textbf{x}_{i}^j-\overline{\textbf{x}}_{i})\|}
+(N-|K^{(t+1)}|)\epsilon\\
=&\sum\limits_{(i,j)\in K^{(r')}}
{\|(\textbf{W}^{(r')})^T(\textbf{x}_{i}^j-\overline{\textbf{x}}_{i})\|}\\
&+\sum\limits_{(i,j)\in {K^{(t+1)}}\backslash K^{(r')}}
{\|(\textbf{W}^{(t+1)})^T(\textbf{x}_{i}^j-\overline{\textbf{x}}_{i})\|}
+(N-|K^{(t+1)}|)\epsilon\\
\leq&\sum\limits_{(i,j)\in K^{(r')}}
{\|(\textbf{W}^{(r')})^T(\textbf{x}_{i}^j-\overline{\textbf{x}}_{i})\|}
+(|K^{(t+1)}|-|K^{(r')}|)\epsilon\\
&+(N-|K^{(t+1)}|)\epsilon\\
=&\sum\limits_{(i,j)\in K^{(r')}}
{\|(\textbf{W}^{(r')})^T(\textbf{x}_{i}^j-\overline{\textbf{x}}_{i})\|}
+(N-|K^{(r')}|)\epsilon
\end{split}
\end{equation}
In addition, \eqref{FfixJineq8} is clearly true for $K^{(r')}=\emptyset$ and $K^{(r')}=K^{(r)}$. Therefore, \eqref{FfixJineq7} holds.
Taking $r=t$ in \eqref{FfixJineq7}, and combing \eqref{FfixJineq6}, it follows
\begin{equation}\label{objineq}
\begin{split}
&\sum\limits_{(i,j)\in K^{(t+1)}} {\|(\textbf{W}^{(t+1)})^T(\textbf{x}_{i}^j-\overline{\textbf{x}}_{i})\|}+(N-|K^{(t+1)}|)\epsilon\\
\leq& \sum\limits_{(i,j)\in K^{(t)}}
{\|(\textbf{W}^{(t)})^T(\textbf{x}_{i}^j-\overline{\textbf{x}}_{i})\|}+(N-K^{(t)}|)\epsilon,
\end{split}
\end{equation}
which is just
\begin{equation}\label{CLDAobj2}
\begin{split}
&\sum\limits_{i=1}^{c}\sum\limits_{j=1}^{N_i}\min(\|(\textbf{W}^{(t+1)})^T(\textbf{x}_{i}^j-\overline{\textbf{x}}_{i})\|,\epsilon)\\
\leq&\sum\limits_{i=1}^{c}\sum\limits_{j=1}^{N_i}\min(\|(\textbf{W}^{(t+1)})^T(\textbf{x}_{i}^j-\overline{\textbf{x}}_{i})\|,\epsilon).
\end{split}
\end{equation}
Therefore, Algorithm 1 monotonically decreases the objective of \eqref{CLDA} in each step. Since the objective of \eqref{CLDA} is obviously lower bounded by 0, this completes the proof.
\hfill $\square$

\section{Experiments}\label{secExp}
In this section, we evaluate the performance of the proposed CLDA compared with its related methods, including LDA \cite{Fisher36, Fukunaga90}, RDA \cite{RLDA}, RLDA \cite{LiShao20}, LDA-L1 \cite{ZhongZhang13,WangLu14}, 
and L2BLDA \cite{LiShaoWang19}. The regularization parameter for RDA is selected from the set $\{10^{-5}, 10^{-4},\ldots,10^{-1},1,10\}$, $\rho$ for RLDA is selected from $\{0.1,0.5,1,5,50,100,1000\}$, the learning rate parameter for LDA-L1 is chosen from $\{10^{-6}, 10^{-5}, \ldots, 10^{2}\}$.
All methods are carried out on a laptop with Intel i5 processor (2.60 GHz) with 4 GB RAM memory using MATLAB 2017a. Experiments are conducted on an artificial data set, some benchmark UCI data sets and two image data sets. Test classification accuracy ($\%$) is used to compare performance, which is obtained by applying nearest neighbor classifier on the projected test data after training dimensionality reduction methods.
\subsection{An artificial data set}
\begin{figure*}[htpb]
\begin{center}{
\subfigure[Training data and projection directions]{
\resizebox*{6.5cm}{!}
{\includegraphics{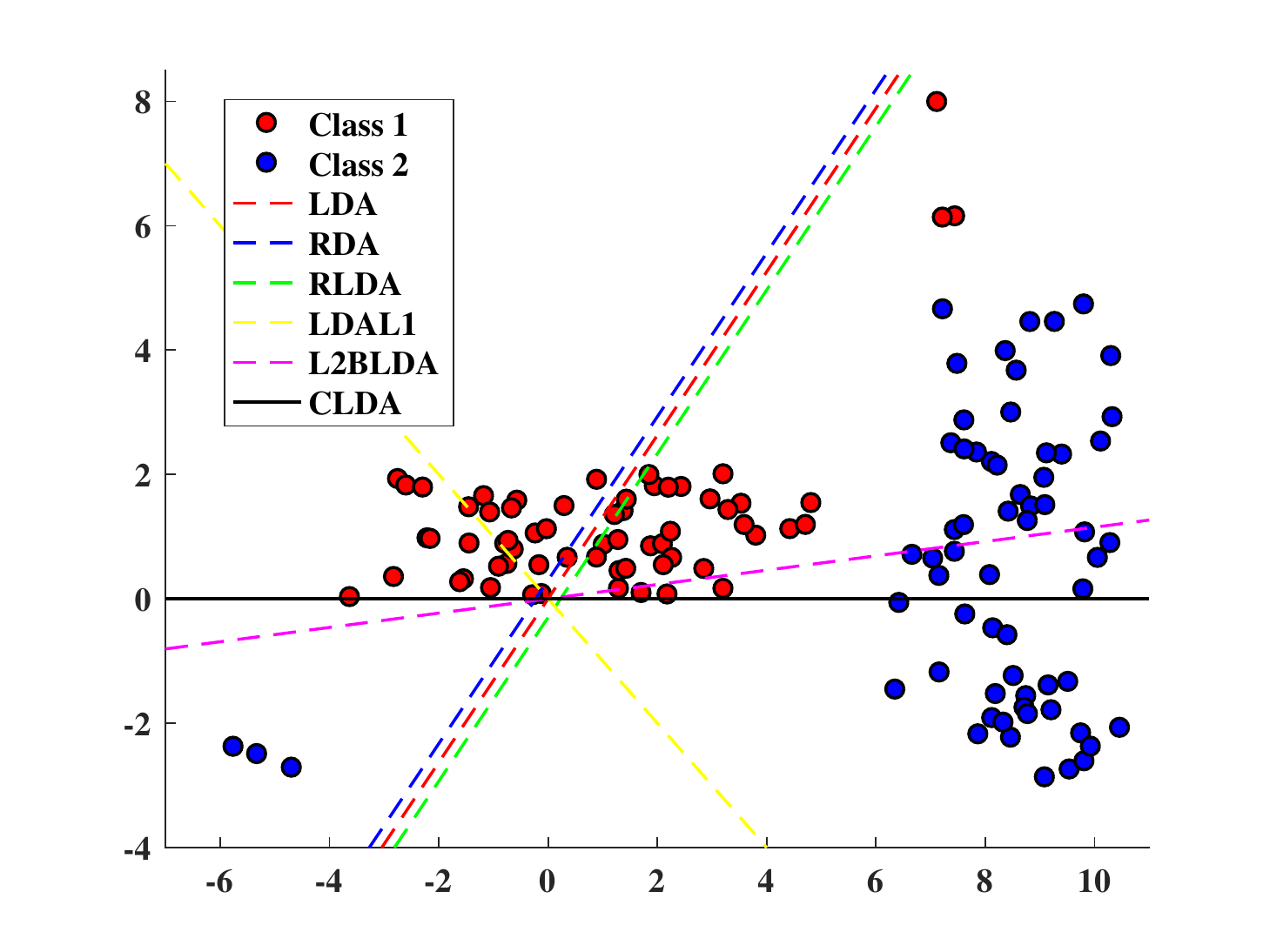}}}\hspace{5pt}
\subfigure[Test data and projection directions]{
\resizebox*{6.5cm}{!}
{\includegraphics{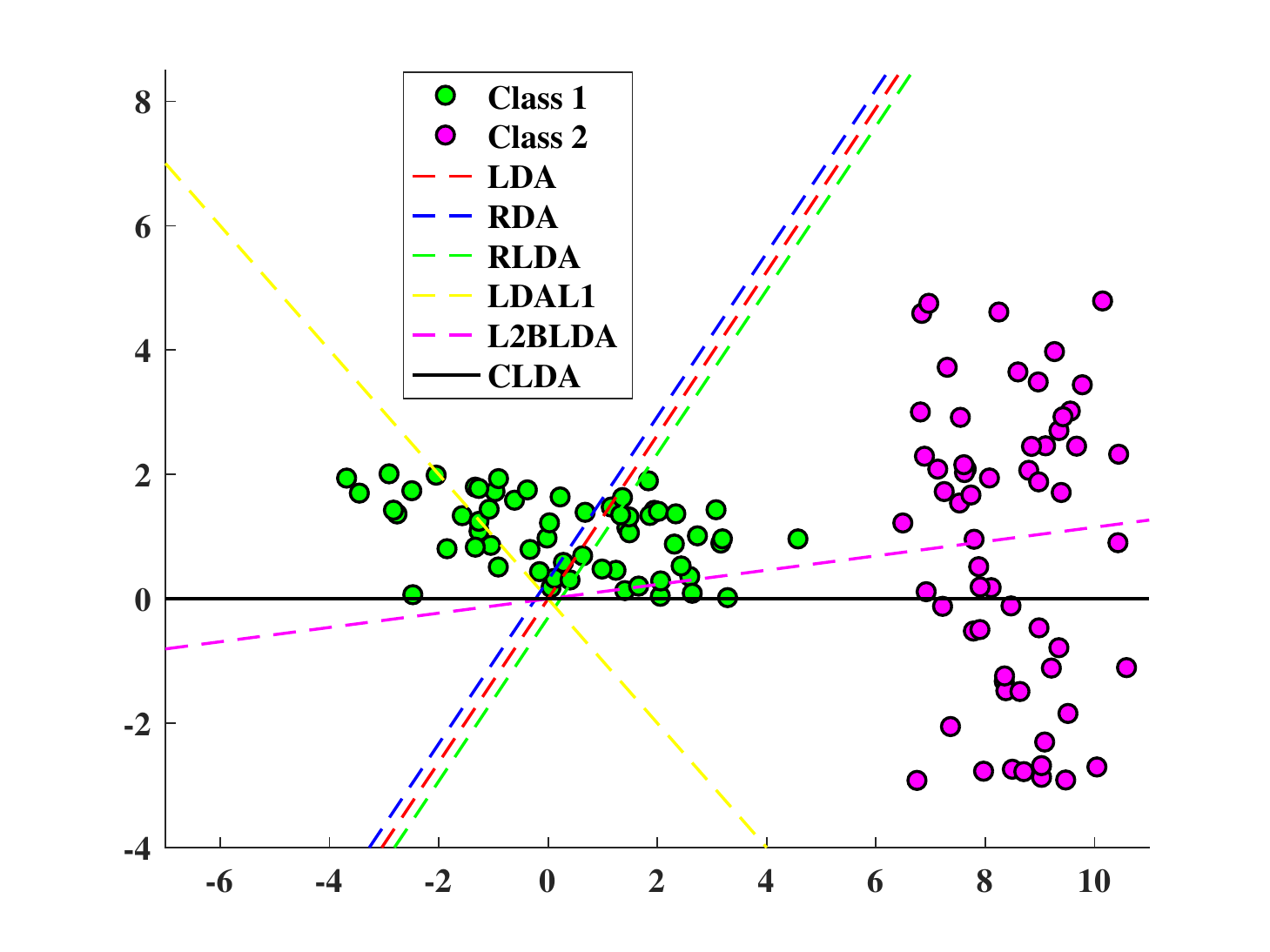}}}\hspace{5pt}
\caption{Training and test artificial data and projection directions obtained by all methods.}
\label{Figarti}}
\end{center}
\end{figure*}

We first consider a two-class two-dimensional artificial data set, and project it to a one-dimensional space.
Class 1 of the data is generated from a uniform distribution that  horizontally distributed with 120 samples, and Class 2 is generated from a uniform distribution that vertically distributed with 120 samples. 50\% samples of each class are randomly selected for training, and the rest ones are used for testing. To test the robustness of the proposed method, we add three extra outliers for each class, as shown in Fig.\ref{Figarti}(a). Test data are shown in Fig.\ref{Figarti}(b).

\begin{figure*}[htpb]
\begin{center}{
\subfigure[LDA]{
\resizebox*{6.5cm}{!}
{\includegraphics{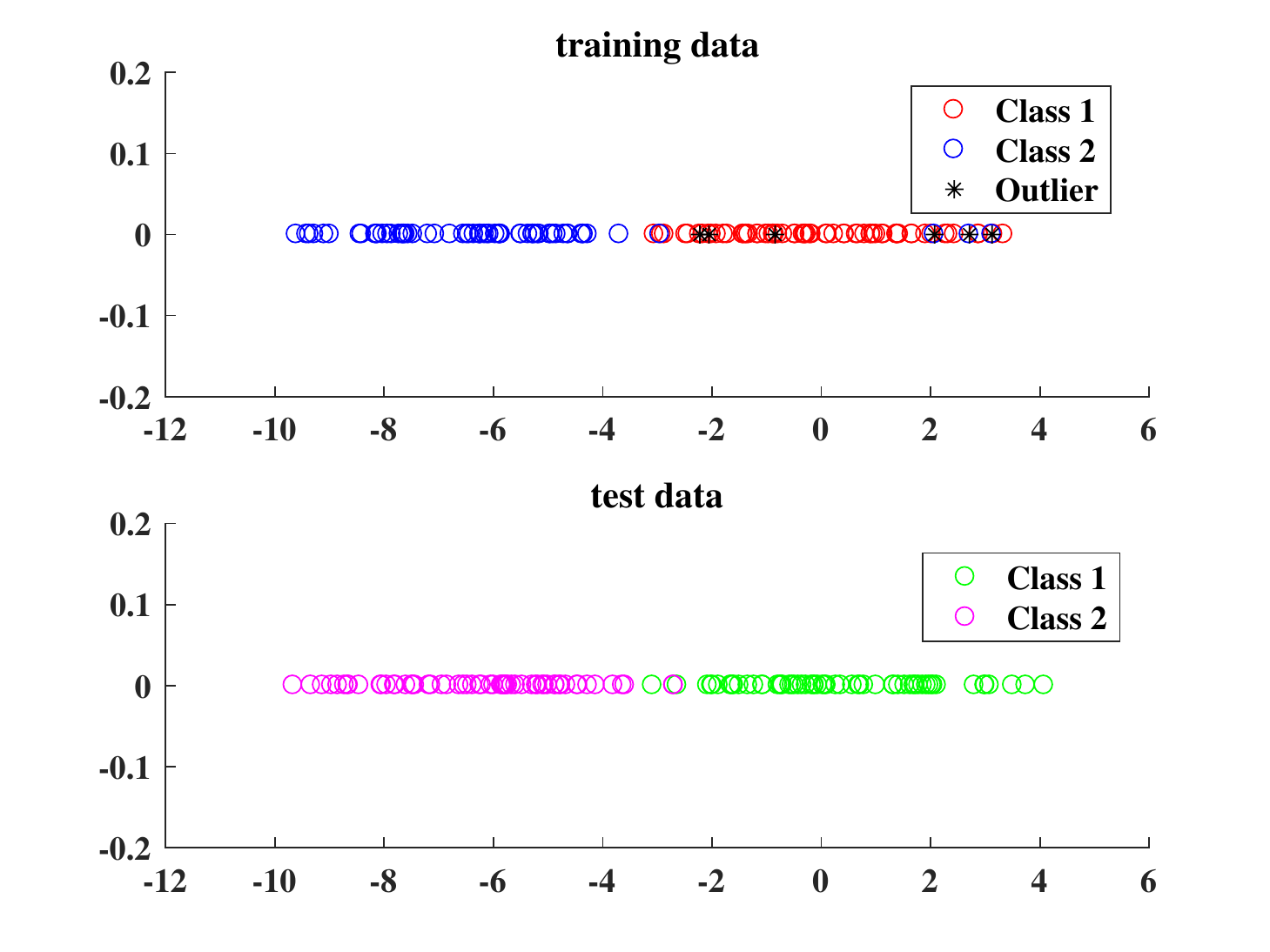}}}\hspace{5pt}
\subfigure[RDA]{
\resizebox*{6.5cm}{!}
{\includegraphics{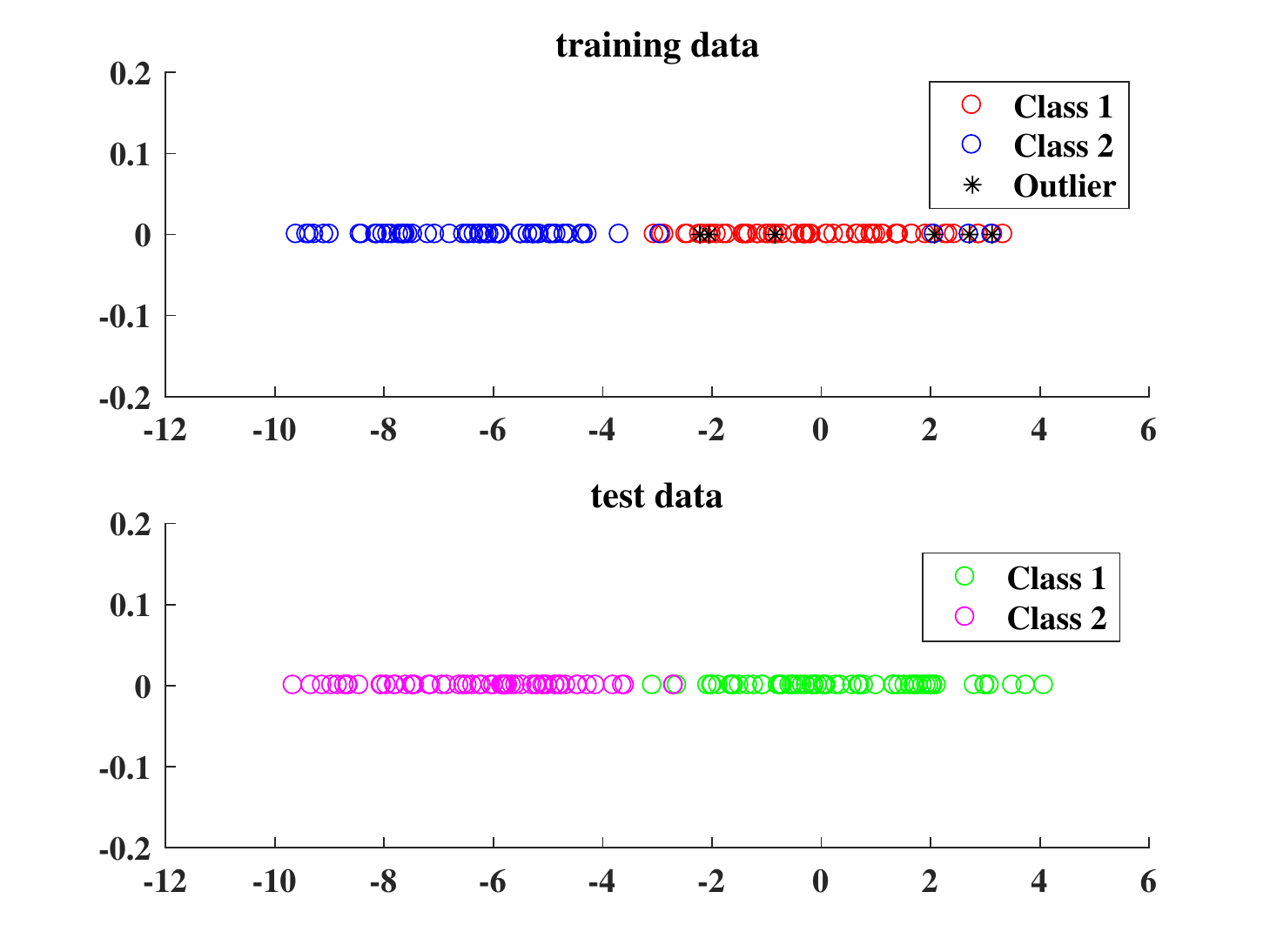}}}\hspace{5pt}
\subfigure[RLDA]{
\resizebox*{6.5cm}{!}
{\includegraphics{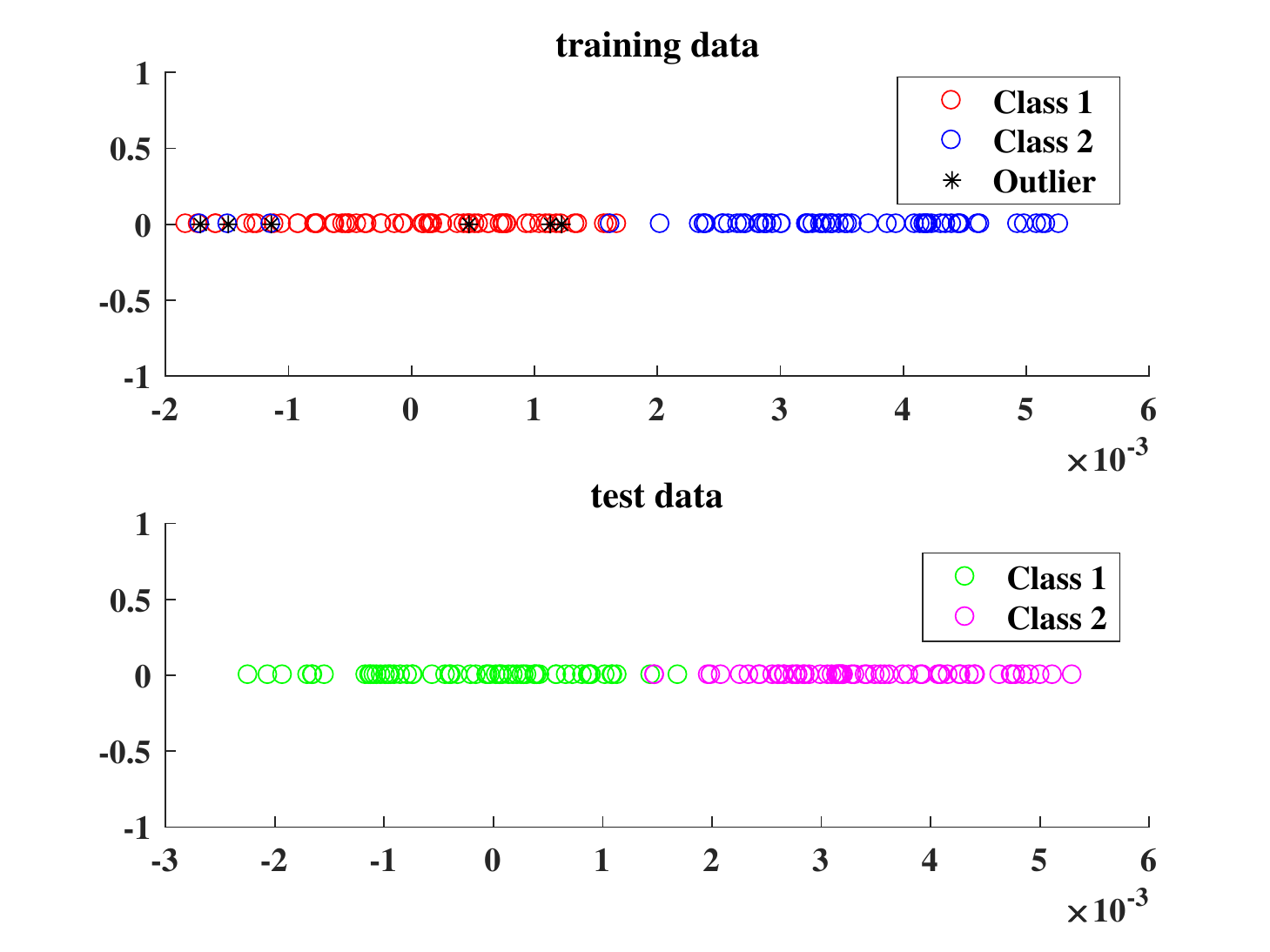}}}\hspace{5pt}
\subfigure[LDAL1]{
\resizebox*{6.5cm}{!}
{\includegraphics{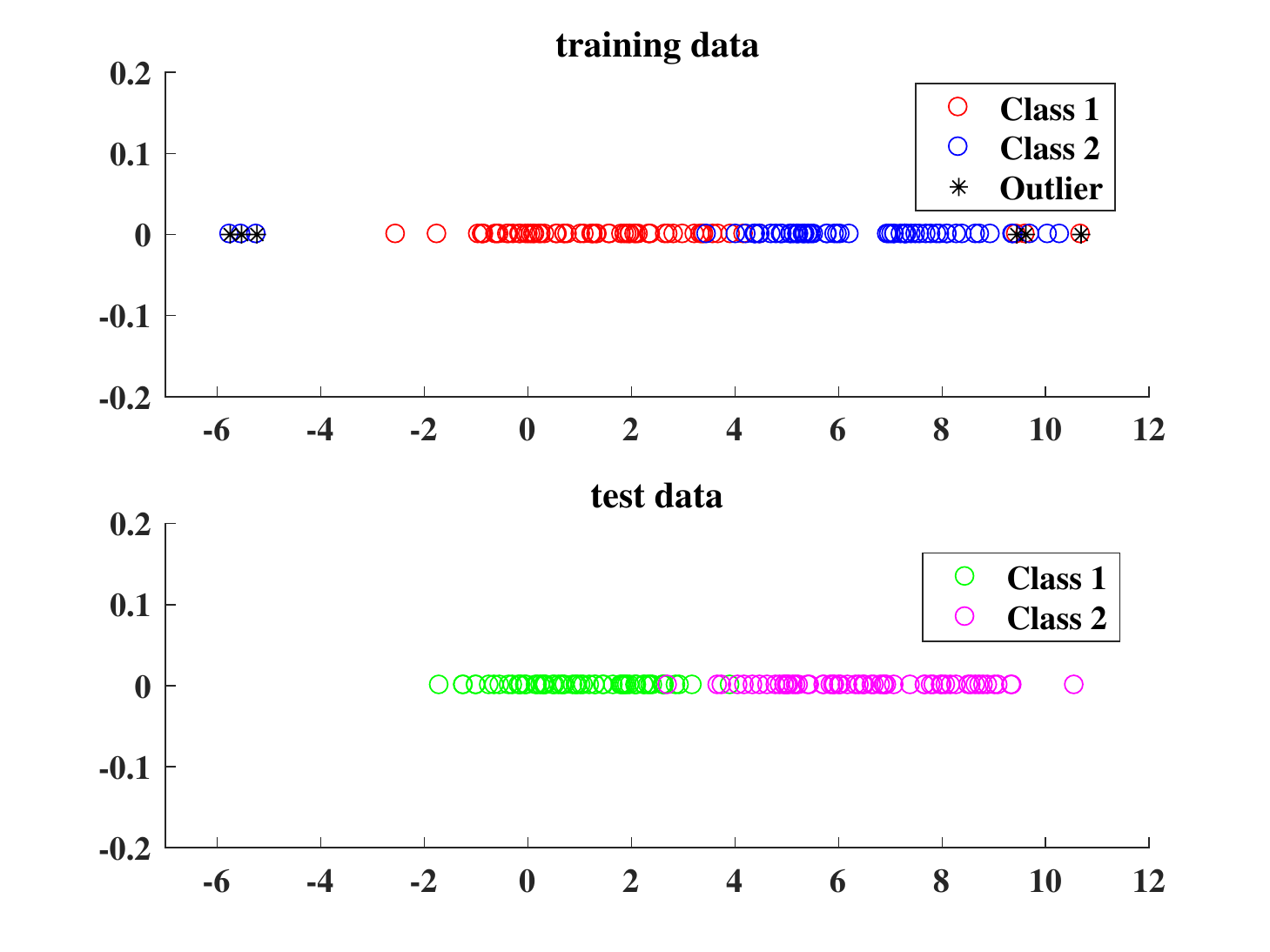}}}\hspace{5pt}
\subfigure[L2BLDA]{
\resizebox*{6.5cm}{!}
{\includegraphics{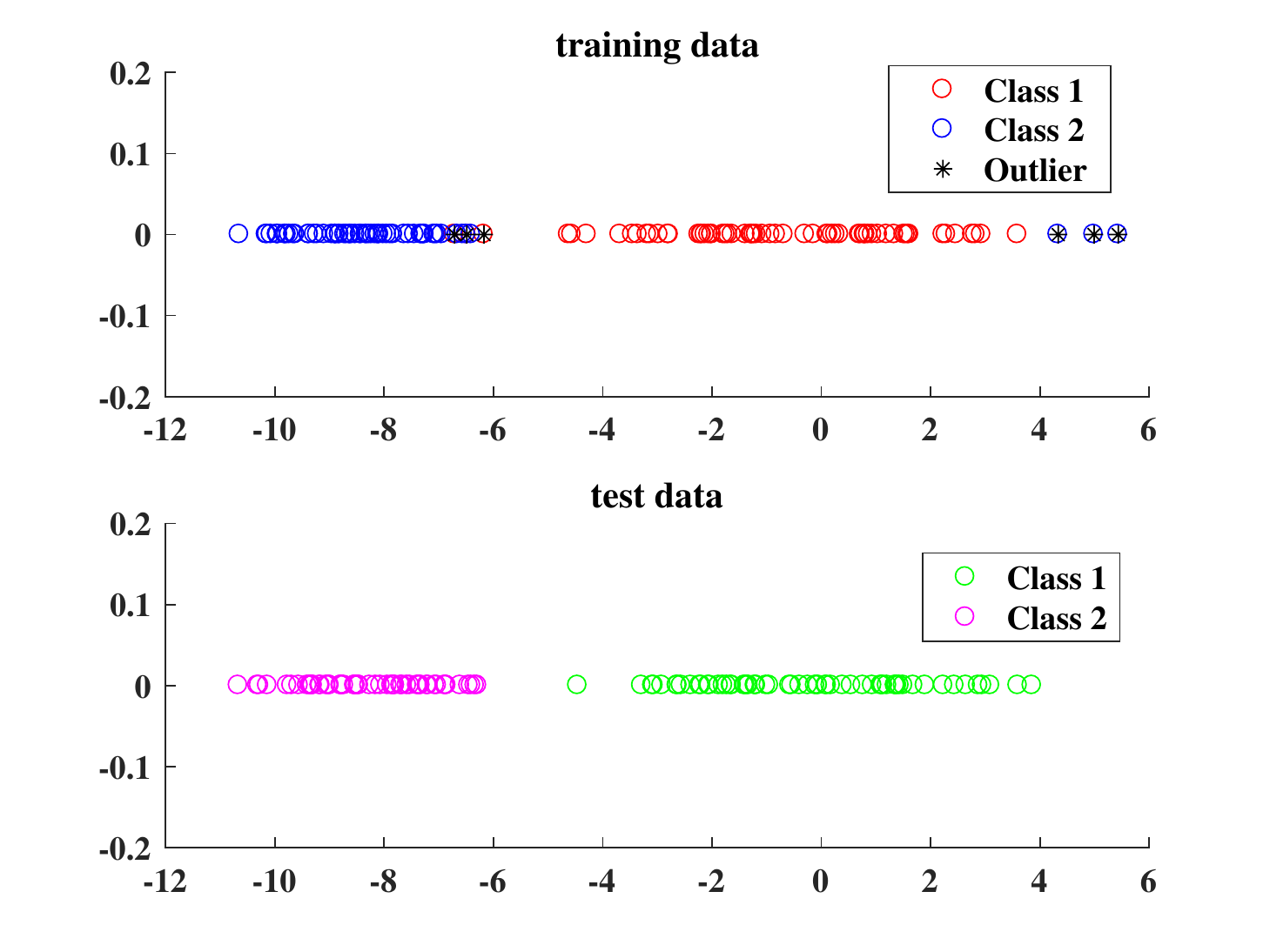}}}\hspace{5pt}
\subfigure[CLDA]{
\resizebox*{6.5cm}{!}
{\includegraphics{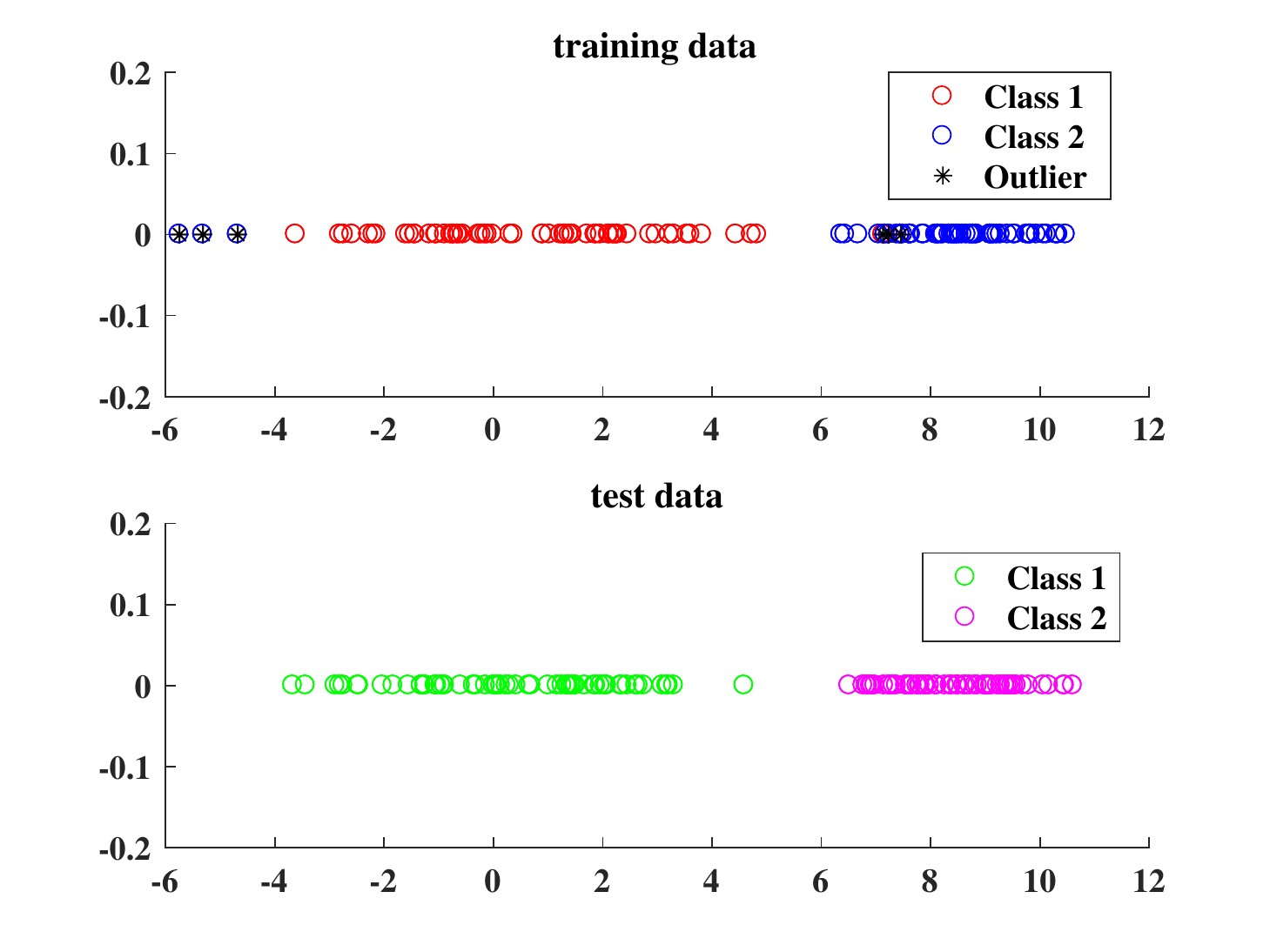}}}\hspace{5pt}
\caption{Training and test artificial data and projection directions obtained by all methods.}
\label{FigProarti}}
\end{center}
\end{figure*}

We apply all methods to the above polluted training data, and use the obtained projection directions to project both training data and test data to one-dimensional space. The classification results on projected test data are shown in Table \ref{Tablearti}, which demonstrates that L2BLDA and our CLDA can separate two classes well. To see the results more clearly, we also show the obtained projection directions.
From the construction of the data, we see the ideal projection direction is parallel to $x$-axis. We plot the projection direction of each method in Fig.\ref{Figarti}.
From the figure, we see the proposed CLDA obtains projection direction that is very close to the ideal one, while other methods have deviation more or less. By further observing the projection data obtained by each method in Fig.\ref{FigProarti}, it can be seen that except L2BLDA and our CLDA, other methods will misclassify samples.
\begin{table}[htbp]
\begin{center}
\caption{Classification results on artificial test data.}
\resizebox{3.3in}{!}
{
\begin{tabular}{l|ccccccccccc}
\toprule
Method & ~LDA~~&RDA~~&RLDA~~&LDA-L1~~&L2BLDA~~&CLDA\\
\midrule
Accuracy~ & 95.00 & 95.00 & 95.00 & 95.83 & \textbf{100.00} & \textbf{100.00} \\
\bottomrule
\end{tabular}
}
\label{Tablearti}
\end{center}
\end{table}

\subsection{Benchmark UCI data sets}

\begin{table}[htbp]
\begin{center}
\renewcommand\tabcolsep{1pt}
\caption{UCI data sets information.}
\resizebox{2.8in}{!}
{
\begin{tabular}{l|ccccclccccc}
\toprule
{\multirow{1}{*}{Data set}} &Sample no.&Feature no.&Class no.\\
\midrule
Australian&  690&14&2\\
BUPA&  345&6&2\\
Car&  1782&6&4\\
Credit&  690&15&2\\
Diabetics&  768&8&2\\
Echo&  131&10&2\\
WPBC& 198&34&2\\
German&  1000&20&2\\
Haberman&  306&3&2\\
Waveform& 5000&21&2\\
House$\_$votes &  435&16&2\\
Iris& 150&4&3\\
Monks3& 432&6&2\\
Sonar& 208&60&2\\
Spect& 267&44&2\\
CMC &1473&9&2\\
Dermatology &366&34&6\\
Glass  &214&9&6\\
Heartc &303&14&2\\
Ionosphere &351&34&2\\
Seeds &  300&2&2\\
\bottomrule
\end{tabular}
}
\label{TableUCIInfo}
\end{center}
\end{table}

In this subsection, the proposed CLDA and its compared methods are applied on 21 benchmark data sets, whose information is listed in Table \ref{TableUCIInfo}. In our experiment, all data are normalized to $[0,1]$. 10-fold cross validation is used for searching optimal parameter, and 10-time average test classification accuracy is adopted. Classification accuracies along with standard deviations for all methods are listed in Table \ref{TableUCIACori}. ``Acc" is short for accuracy ($\%$), and ``Std" is short for standard deviation.
By observing the results in Table \ref{TableUCIACori}, we see that the proposed CLDA outperforms other methods on most of the data sets. On other data sets, CLDA performs comparable to the one with highest accuracy for most cases. To clearly see the superiority of CLDA, we compute the rank of each methods, as shown in Table \ref{TableUCIRankori}. It can be seen that on most data sets CLDA has the highest rank, and on most other data sets its rank is also on the front. This leads to its highest average rank over all data sets.

\begin{table*}[htbp]
\begin{center}
\caption{Classification results on original UCI data sets.}
\resizebox{5.2in}{!}
{
\begin{tabular}{l|ccccccccccc}
\toprule
{\multirow{2}{*}{Data set}} & LDA &RDA&RLDA&LDA-L1~&L2BLDA&CLDA\\
& Acc$\pm$Std~~& Acc$\pm$Std~~& Acc$\pm$Std~~& Acc$\pm$Std~~& Acc$\pm$Std~~&Acc$\pm$Std\\
\midrule
Australian & 82.14 $\pm$ 4.62 & 81.95 $\pm$ 4.85& 83.74 $\pm$ 4.02& \textbf{83.76 $\pm$ 3.31}&  \textbf{83.76 $\pm$ 3.31}& 82.02 $\pm$ 5.46\\
BUPA & 57.66 $\pm$ 4.53 & 56.21 $\pm$ 8.73& 66.03 $\pm$ 7.30& 67.39 $\pm$ 4.13& 67.39 $\pm$ 4.13& \textbf{69.13 $\pm$ 5.99}\\
Car & 84.21 $\pm$ 4.84 & 84.86 $\pm$ 4.28& 85.91 $\pm$ 7.85& 94.34 $\pm$ 6.14& \textbf{ 95.43 $\pm$ 2.69}& 94.04 $\pm$ 4.17\\
Credit & 81.46 $\pm$ 4.02 & 81.15 $\pm$ 4.94& 83.44 $\pm$ 4.02& \textbf{83.69 $\pm$ 4.84}&  \textbf{83.69 $\pm$ 4.84}& 83.30 $\pm$ 4.03\\
Diabetics & 67.35 $\pm$ 4.61 & 70.18 $\pm$ 4.15& 73.38 $\pm$ 3.53& 71.31 $\pm$ 4.30& \textbf{76.67 $\pm$ 3.79}& 73.80 $\pm$ 5.64\\
Echo & 89.34 $\pm$ 8.97 & 87.19 $\pm$ 10.56& 87.56 $\pm$ 12.18& 91.59 $\pm$ 8.47&  91.59 $\pm$ 8.47& \textbf{97.02 $\pm$ 3.86}\\
WPBC & 78.39 $\pm$ 8.17 & 76.53 $\pm$ 13.88&\textbf{ 82.57 $\pm$ 8.09}& 80.95 $\pm$ 7.08&  80.95 $\pm$ 7.08& 81.55 $\pm$ 9.49\\
German & 67.50 $\pm$ 3.69 & 68.60 $\pm$ 3.66& 72.90 $\pm$ 2.60& 73.80 $\pm$ 4.21&  \textbf{75.80 $\pm$ 2.66}& 75.40 $\pm$ 3.95\\
Haberman & 63.87 $\pm$ 11.70 & 67.46 $\pm$ 9.16& 62.18 $\pm$ 10.46& 67.53 $\pm$ 5.03& 67.53 $\pm$ 5.03&\textbf{69.94 $\pm$ 6.75}\\
Waveform & 79.99 $\pm$ 1.45 & 80.23 $\pm$ 1.78& 80.93 $\pm$ 2.19& \textbf{87.12 $\pm$ 1.18}& \textbf{87.12 $\pm$ 1.18}& 83.94 $\pm$ 1.62\\
House$\_$votes & 93.10 $\pm$ 1.87 & 93.60 $\pm$ 3.00& 94.50 $\pm$ 3.62& \textbf{95.38 $\pm$ 2.68}& \textbf{ 95.38 $\pm$ 2.68}& 95.19 $\pm$ 2.26\\
Iris & 96.00 $\pm$ 3.44 & 96.00 $\pm$ 3.44& 94.67 $\pm$ 6.89& \textbf{96.67 $\pm$ 5.67}& 94.00 $\pm$ 4.92& \textbf{96.67 $\pm$ 4.71}\\
Monks3 & 69.21 $\pm$ 13.83 & 61.34 $\pm$ 10.95& 68.88 $\pm$ 4.84& 70.34 $\pm$ 8.82& 70.34 $\pm$ 8.82& \textbf{84.02 $\pm$ 12.53}\\
Sonar & 71.69 $\pm$ 12.08 & 73.63 $\pm$ 10.29& 89.51 $\pm$ 8.36& 89.54 $\pm$ 6.79&  90.82 $\pm$ 5.64& \textbf{91.90 $\pm$ 5.52}\\
Spect & 72.89 $\pm$ 9.25 & 69.32 $\pm$ 9.45& 81.95 $\pm$ 6.45& 84.84 $\pm$ 6.09&  84.84 $\pm$ 6.09& \textbf{88.01 $\pm$ 5.56}\\
CMC & 68.91 $\pm$ 3.66 & 72.85 $\pm$ 2.26& 69.17 $\pm$ 3.43& 70.60 $\pm$ 2.17&  72.93 $\pm$ 2.00& \textbf{74.51 $\pm$ 2.49}\\
Dermatology & 97.26 $\pm$ 2.60 & 96.99 $\pm$ 3.03& 96.49 $\pm$ 2.84& 97.51 $\pm$ 3.08&  94.44 $\pm$ 4.14& \textbf{97.83 $\pm$ 1.74}\\
Glass & 64.68 $\pm$ 11.64 & 71.84 $\pm$ 11.28& 69.32 $\pm$ 10.14& 69.69 $\pm$ 7.70&  64.62 $\pm$ 11.40& \textbf{82.56 $\pm$ 8.65}\\
Heartc & 96.32 $\pm$ 3.34 & 89.17 $\pm$ 7.41& 88.66 $\pm$ 31.23& 92.73 $\pm$ 5.41&  88.50 $\pm$ 6.10& \textbf{96.40 $\pm$ 2.90}\\
Ionosphere & 81.60 $\pm$ 8.63 & 94.02 $\pm$ 3.42& 84.86 $\pm$ 30.03& 91.20 $\pm$ 4.99& 93.44 $\pm$ 2.72& \textbf{94.06 $\pm$ 3.33}\\
Seeds & 96.19 $\pm$ 4.38 & 95.71 $\pm$ 2.70& \textbf{97.14 $\pm$ 4.02}& 93.81 $\pm$ 5.04& 92.86 $\pm$ 4.63& 94.29 $\pm$ 5.41\\
\bottomrule
\end{tabular}
}
\label{TableUCIACori}
\end{center}
\end{table*}

\begin{table}[htbp]
\begin{center}
\caption{Classification ranks on original UCI data sets.}
\resizebox{3.3in}{!}
{
\begin{tabular}{l|ccccccccccc}
\toprule
{\multirow{2}{*}{Data set}} & LDA& RDA &RLDA~~&LDA-L1~~&L2BLDA&CLDA\\
& Rank& Rank& Rank& Rank& Rank& Rank\\
\midrule
Australian  & 4 & 6 & 3 & 1.5 & 1.5 & 5  \\
BUPA  & 5 & 6 & 4 & 2.5 & 2.5 & 1  \\
Car  & 6 & 5 & 4 & 2 & 1 & 3  \\
Credit  & 5 & 6 & 3 & 1.5 & 1.5 & 4  \\
Diabetics  & 6 & 5 & 3 & 4 & 1 & 2  \\
Echo  & 4 & 6 & 5 & 2.5 & 2.5 & 1  \\
WPBC  & 5 & 6 & 1 & 3.5 & 3.5 & 2  \\
German  & 6 & 5 & 4 & 3 & 1 & 2  \\
Haberman  & 5 & 4 & 6 & 2.5 & 2.5 & 1  \\
Waveform  & 6 & 5 & 4 & 1.5 & 1.5 & 3  \\
House$\_$votes  & 6 & 5 & 4 & 1.5 & 1.5 & 3  \\
Iris  & 3.5 & 3.5 & 5 & 1.5 & 6 & 1.5  \\
Monks3  & 4 & 6 & 5 & 2.5 & 2.5 & 1  \\
Sonar  & 6 & 5 & 4 & 3 & 2 & 1  \\
Spect  & 5 & 6 & 4 & 2.5 & 2.5 & 1  \\
CMC  & 6 & 3 & 5 & 4 & 2 & 1  \\
Dermatology  & 3 & 4 & 5 & 2 & 6 & 1  \\
Glass  & 5 & 2 & 4 & 3 & 6 & 1  \\
Heartc  & 2 & 4 & 5 & 3 & 6 & 1  \\
Ionosphere  & 6 & 2 & 5 & 4 & 3 & 1  \\
Seeds  & 2 & 3 & 1 & 5 & 6 & 4  \\
\midrule
Average rank& 4.7857~ & 4.6429~ & 4.0000 & 2.6905 & 2.9524 & \textbf{1.9286} \\
\bottomrule
\end{tabular}
}
\label{TableUCIRankori}
\end{center}
\end{table}

To test the robustness of the proposed CLDA, we consider noise polluted data. Specifically, random 30\% features of random 10\% percent samples are selected and added with random Gaussian noise of mean zero and variance 0.05, respectively. The classification results on noise data are listed in Table \ref{TableUCIACnoise}.
From the table, we have the following observations: (i) The performance for all methods degenerates on most data sets. (ii) Robustness designed methods generally perform better than L2-norm ones. In specific, RLDA, LDA-L1, L2BLDA and CLDA perform much better than LDA and RDA; (iii) The proposed CLDA is less affected by noise comparing to other methods. (iv) The rank results demonstrated in Table \ref{TableUCIRanknoise} support the advantage of CLDA. In fact, the average rank of CLDA becomes higher on noise data comparing to its average rank on original data.

\begin{table*}[htbp]
\begin{center}
\caption{Classification results on noise UCI data sets.}
\resizebox{5.2in}{!}
{
\begin{tabular}{l|ccccccccccc}
\toprule
{\multirow{2}{*}{Data set}} & LDA &RDA&RLDA&LDA-L1&L2BLDA&CLDA\\
& Acc $\pm$ Std~~& Acc $\pm$ Std~~& Acc $\pm$ Std~~& Acc $\pm$ Std~~& Acc $\pm$ Std~~&Acc $\pm$ Std\\
\midrule
Australian & 77.81 $\pm$ 6.40 & 81.46 $\pm$ 3.32& \textbf{85.39 $\pm$ 4.20}& 82.66 $\pm$ 3.57&  83.35 $\pm$ 4.54& 80.86$\pm$ 5.66\\
BUPA & 54.30 $\pm$ 9.69 & 55.45 $\pm$ 11.93& 61.07 $\pm$ 6.10& 63.24 $\pm$ 5.76&  66.96 $\pm$ 5.00& \textbf{67.25 $\pm$ 7.45}\\
Car & 83.96 $\pm$ 3.66 & 87.78 $\pm$ 4.29& 84.49 $\pm$ 5.77& \textbf{94.55 $\pm$ 1.22}& 88.34 $\pm$ 2.84& 94.01 $\pm$ 1.81\\
Credit & 83.08 $\pm$ 5.47 & 77.36 $\pm$ 5.21& 83.71 $\pm$ 3.03& 84.68 $\pm$ 4.52&  85.63 $\pm$ 4.63& \textbf{86.23 $\pm$ 2.10}\\
Diabetics & 64.55 $\pm$ 4.50 & 67.11 $\pm$ 4.49& 69.63 $\pm$ 4.63& 73.08 $\pm$ 4.19&  68.84 $\pm$ 5.07& \textbf{73.56 $\pm$ 5.32}\\
Echo & 86.25 $\pm$ 11.75 & 84.95 $\pm$ 8.13& 88.85 $\pm$ 9.27& 89.45 $\pm$ 7.69&  86.29 $\pm$ 7.00& \textbf{93.89 $\pm$ 5.91}\\
WPBC & 71.13 $\pm$ 6.91 & 69.94 $\pm$ 8.13& 81.34 $\pm$ 6.17& 83.03 $\pm$ 4.78& \textbf{82.14 $\pm$ 6.89}& 81.68 $\pm$ 7.11\\
German & 70.00 $\pm$ 4.59 & 69.00 $\pm$ 4.59& 74.90 $\pm$ 2.33& 72.20 $\pm$ 2.04&  71.40 $\pm$ 3.17& \textbf{75.10 $\pm$ 2.88}\\
Haberman & 62.82 $\pm$ 9.32 & 64.32 $\pm$ 7.61& 67.20 $\pm$ 3.47& 68.85 $\pm$ 3.58&  68.61 $\pm$ 7.85& \textbf{71.08 $\pm$ 4.60}\\
Waveform & 79.48 $\pm$ 1.54 & 80.32 $\pm$ 1.37& 81.78 $\pm$ 1.62& \textbf{87.28 $\pm$ 1.42}&  86.33 $\pm$ 1.00& 83.94 $\pm$ 1.35\\
House$\_$votes & 95.40 $\pm$ 3.06 & 93.52 $\pm$ 4.46& 94.94 $\pm$ 3.41& 95.63 $\pm$ 2.27&  93.81 $\pm$ 2.15& \textbf{95.65 $\pm$ 2.96}\\
Iris & 98.67 $\pm$ 2.81 & 98.00 $\pm$ 3.22& 92.00 $\pm$ 8.20& 94.67 $\pm$ 6.13&  93.33 $\pm$ 6.29& \textbf{98.67 $\pm$ 2.81}\\
Monks3 & 61.60 $\pm$ 9.78 & 64.54 $\pm$ 15.80& 80.63 $\pm$ 13.39& 64.54 $\pm$ 13.39&  76.48 $\pm$ 6.63& \textbf{81.79 $\pm$ 7.23}\\
Sonar & 77.10 $\pm$ 8.70 & 69.12 $\pm$ 6.01& 89.85 $\pm$ 4.05& 88.64 $\pm$ 6.40& 90.83 $\pm$ 4.10& \textbf{91.83 $\pm$ 3.92}\\
Spect & 70.38 $\pm$ 6.88 & 73.81 $\pm$ 8.20& 83.53 $\pm$ 7.77& 86.48 $\pm$ 7.21&  83.11 $\pm$ 6.40& \textbf{86.88 $\pm$ 3.25}\\
CMC & 70.29 $\pm$ 2.53 & 69.95 $\pm$ 3.10& \textbf{74.24 $\pm$ 1.99}& 72.89 $\pm$ 2.59&  72.84 $\pm$ 2.44& 74.13 $\pm$ 1.38\\
Dermatology & 97.29 $\pm$ 2.88 & 97.31 $\pm$ 1.78& 94.77 $\pm$ 2.71& \textbf{98.04 $\pm$ 1.88}&  96.16 $\pm$ 3.22& 97.51 $\pm$ 1.33\\
Glass & 69.02 $\pm$ 12.17 & 68.81 $\pm$ 7.11& 70.19 $\pm$ 6.97& 70.83 $\pm$ 7.90&  72.81 $\pm$ 12.04& \textbf{74.88 $\pm$ 10.04}\\
Heartc & 96.41 $\pm$ 2.41 & 95.34 $\pm$ 3.94& 95.08 $\pm$ 4.95& 92.16 $\pm$ 3.76&  88.38 $\pm$ 4.21& \textbf{96.39 $\pm$ 2.74}\\
Ionosphere & 84.56 $\pm$ 5.86 & 82.66 $\pm$ 6.93& 89.43 $\pm$ 5.30& 94.30 $\pm$ 2.70&  92.14 $\pm$ 3.63& \textbf{91.52 $\pm$ 1.98}\\
Seeds & 94.29 $\pm$ 4.92 & 95.24 $\pm$ 3.89& 94.29 $\pm$ 4.38& 92.38 $\pm$ 5.59&  64.54 $\pm$ 5.52& \textbf{95.71 $\pm$ 3.51}\\
\bottomrule
\end{tabular}
}
\label{TableUCIACnoise}
\end{center}
\end{table*}

\begin{table}[htbp]
\begin{center}
\caption{Classification ranks on noise UCI data sets.}
\resizebox{3.3in}{!}
{
\begin{tabular}{l|ccccccccccc}
\toprule
{\multirow{2}{*}{Data set}} & LDA& RDA &RLDA~~&LDA-L1~~&L2BLDA&CLDA\\
& Rank& Rank& Rank& Rank& Rank& Rank\\
\midrule
Australian  & 6 & 4 & 1 & 3 & 2 & 5  \\
BUPA  & 6 & 5 & 4 & 3 & 2 & 1  \\
Car  & 6 & 4 & 5 & 1 & 3 & 2  \\
Credit  & 5 & 6 & 4 & 3 & 2 & 1  \\
Diabetics  & 6 & 5 & 3 & 2 & 4 & 1  \\
Echo  & 5 & 6 & 3 & 2 & 4 & 1  \\
WPBC  & 5 & 6 & 4 & 1 & 2 & 3  \\
German  & 5 & 6 & 2 & 3 & 4 & 1  \\
Haberman  & 6 & 5 & 4 & 2 & 3 & 1  \\
Waveform  & 6 & 5 & 4 & 1 & 2 & 3  \\
House$\_$votes  & 3 & 6 & 4 & 2 & 5 & 1  \\
Iris  & 1.5 & 3 & 6 & 4 & 5 & 1.5  \\
Monks3  & 6 & 4.5 & 2 & 4.5 & 3 & 1  \\
Sonar  & 5 & 6 & 3 & 4 & 2 & 1  \\
Spect  & 6 & 5 & 3 & 2 & 4 & 1  \\
CMC  & 5 & 6 & 1 & 3 & 4 & 2  \\
Dermatology  & 4 & 3 & 6 & 1 & 5 & 2  \\
Glass  & 5 & 6 & 4 & 3 & 2 & 1  \\
Heartc  & 1 & 3 & 4 & 5 & 6 & 2  \\
Ionosphere  & 5 & 6 & 4 & 1 & 2 & 3  \\
Seeds  & 3.5 & 2 & 3.5 & 5 & 6 & 1  \\
\midrule
Average rank~~&
4.8095~ & 4.8810~ & 3.5476 & 2.6429 & 3.4286 & \textbf{1.6905} \\
\bottomrule
\end{tabular}
}
\label{TableUCIRanknoise}
\end{center}
\end{table}

\subsection{Image data sets}

\subsubsection{Coil100 data set}
\begin{figure}[htpb]
\begin{center}{{
\resizebox*{7.0cm}{!}
{\includegraphics{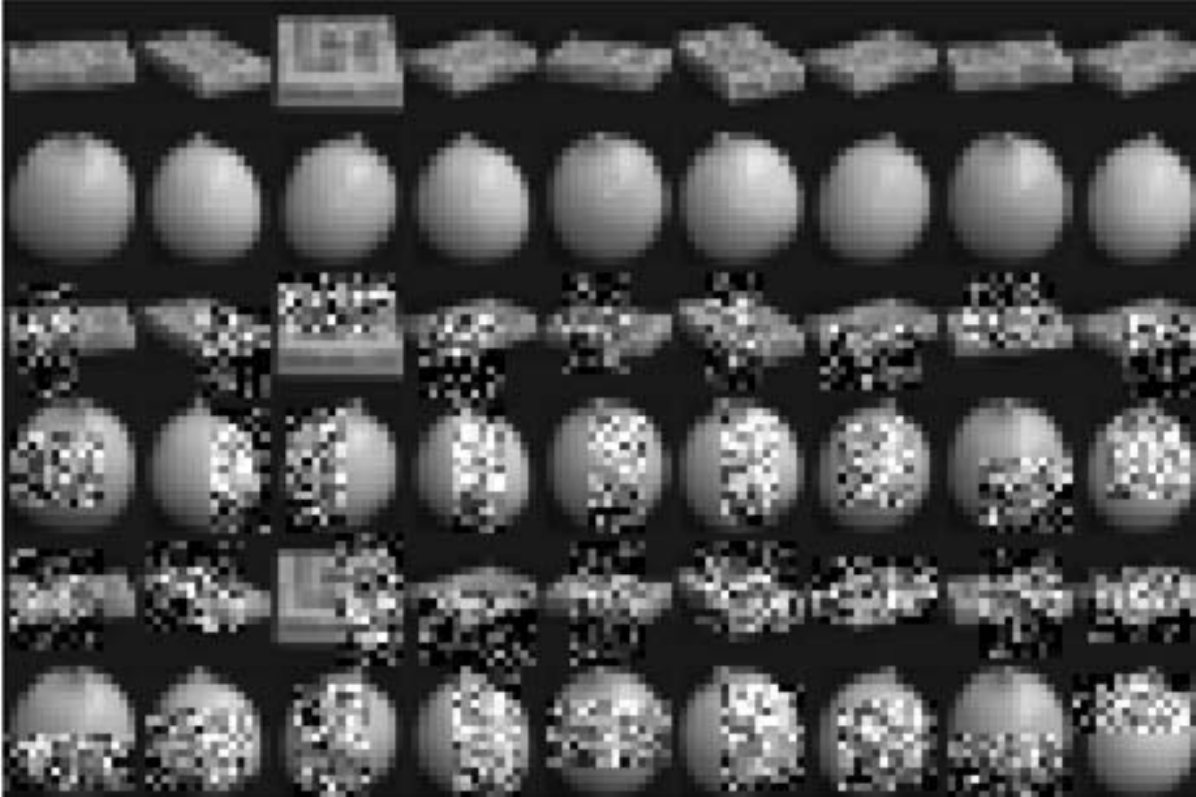}}}\hspace{5pt}
\caption{Original and noise sample images from Coil100 data set.}
\label{Coil100sample}}
\end{center}
\end{figure}

In this subsection, the behaviors of various methods are investigated on two image data sets. The first one is the COIL100 data set that includes 100 image objects. For each object, its images were taken five degrees apart as the object was rotated on a turntable. All images are resized to 16$\times$16 pixel. In our experiments, 50\% images are randomly chosen to form the training set, and the rest images form the test set.
To test the robustness of each method, we not only perform experiments on original data, but also consider polluted training data. In specific, we add random Gaussian noise of mean zero and variance 0.05 that covers 30\% and 40\% rectangular area of each training image, and use the projection matrix obtained on polluted data for testing. Original sample training images and corresponding polluted ones are shown in Fig.\ref{Coil100sample}.

The recognition results of all methods on these data sets are listed in Table \ref{TableCoil}. From the table, we see on original data, the proposed CLDA has comparable performance to LDA-L1 and L2BLDA, and is better than LDA, RDA and RLDA. However, when noise is added, the performance of CLDA is barely affected, while performance of other methods is influenced by noise dramatically. The results show the robustness of CLDA, and it has better discriminant ability on noise data. We also list the ranks of various methods in Table \ref{TableCoilRank}. The highest rank of CLDA on noise data supports its robustness.

\begin{figure*}[htpb]
\begin{center}{
\subfigure[Original data]{
\resizebox*{5.7cm}{!}
{\includegraphics{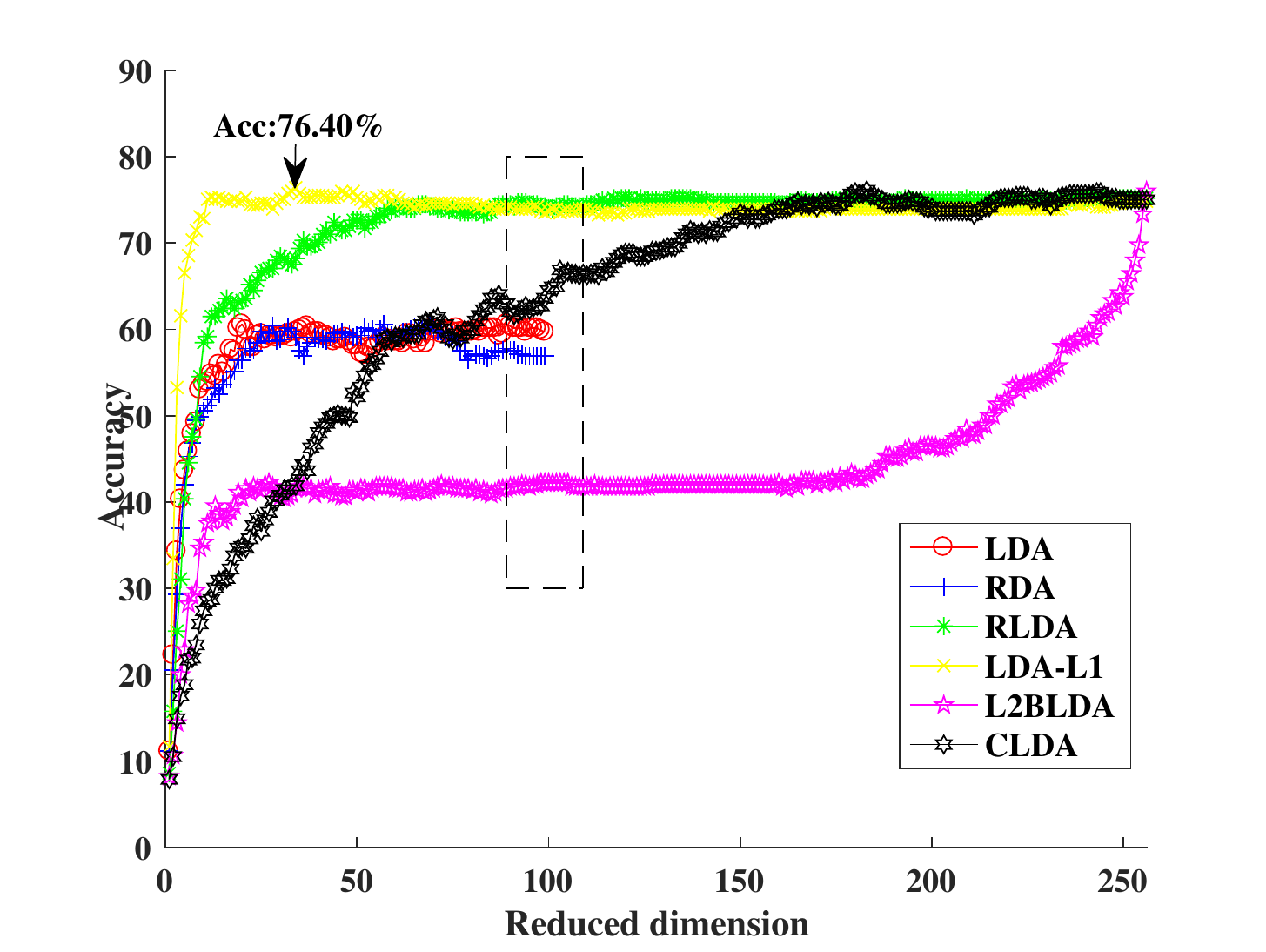}}}\hspace{5pt}
\subfigure[30\% noise data]{
\resizebox*{5.7cm}{!}
{\includegraphics{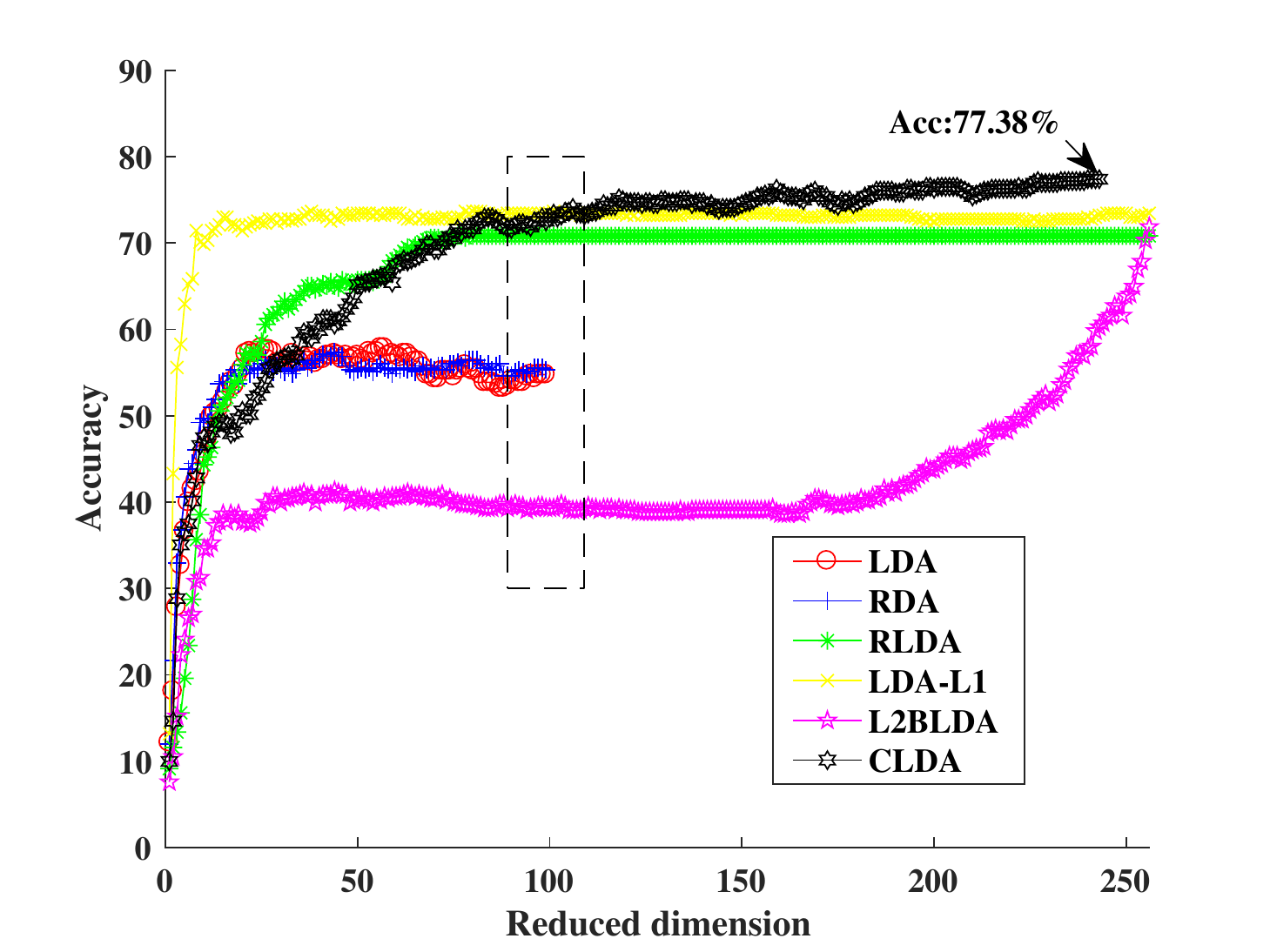}}}\hspace{5pt}
\subfigure[40\% noise data]{
\resizebox*{5.7cm}{!}
{\includegraphics{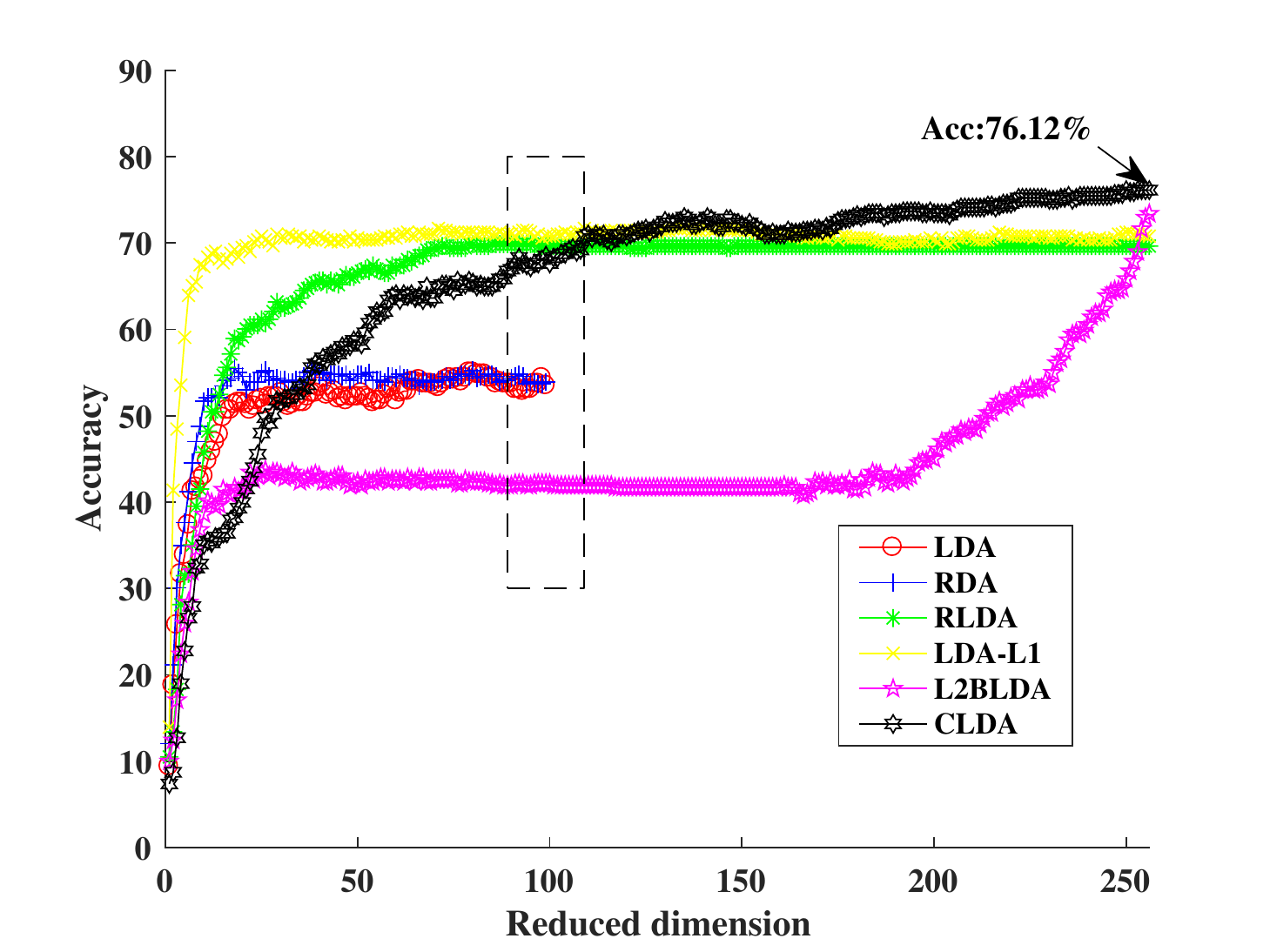}}}\hspace{5pt}
\caption{The variation of accuracies along different dimensions on Coil100 data set.}
\label{Coildim}}
\end{center}
\end{figure*}

To further investigate the behavior of CLDA under different reduced dimensions, Fig.\ref{Coildim} describes the variation of accuracies of all methods along reduced dimensions. The figure shows that in general, as the number of dimensions varies, the accuracies for all methods in general arise. For LDA and RDA, due to their rank limit, their maximum number of reduced dimensions is restricted by $c-1$, where $c$ is the number of class. For Coil100 data set, the maximum number is 99. It should also be noted that for our CLDA, it may also encounter similar situation, since the number of reduced dimensions is decided by the rack of $\textbf{S}_1=\textbf{H}_w\textbf{F}\textbf{H}_w^T$ in \eqref{CLDAobj}. Then rank of $\textbf{S}_1$ is clearly related to $\textbf{H}_w$ and $\textbf{F}$, while $\textbf{F}$ is changing during the solving procedure iteration. Therefore, the rank of $\textbf{S}_1$ is not deterministic. However, through experiments, we find out that CLDA can achieve high reduced dimensions. In fact, CLDA can extract the maximum number of reduced dimensions on original and 40\% noise Coil100 data, and almost achieve maximum number on 30\% noise Coil100 data, which is much lager than LDA and RDA. The results in Fig.\ref{Coildim} show that reduced dimension has a great influence to the behavior of all methods, and it is necessary to choose an optimal one. Under optimal dimensions, the proposed CLDA behaves well.

\begin{table}[htbp]
\begin{center}
\caption{Classification results on Coil100 data.}
\resizebox{3.3in}{!}
{
\begin{tabular}{l|ccccccccccc}
\toprule
Data set & LDA ~&RDA~&RLDA~&LDA-L1~&L2BLDA~&CLDA\\
\midrule
Coil100ori~  &~ 60.58 ~& 60.61 & 75.17 & \textbf{76.40} & 76.00 & 76.10 \\
Coil100g30~  &~ 57.84 ~& 57.34 & 70.82 & 73.66 & 71.88 & \textbf{77.38} \\
Coil100g40~  &~ 55.02 ~& 55.46 & 69.87 & 71.68 & 70.70 & \textbf{76.12} \\
\bottomrule
\end{tabular}
}
\label{TableCoil}
\end{center}
\end{table}

\begin{table}[htbp]
\begin{center}
\caption{Classification ranks on Coil100 data.}
\resizebox{3.3in}{!}
{
\begin{tabular}{l|ccccccccccc}
\toprule
{\multirow{2}{*}{Data set}} & LDA& RDA &RLDA~~&LDA-L1~~&L2BLDA&CLDA\\
& Rank& Rank& Rank& Rank& Rank& Rank\\
\midrule
Coil100ori  & 6.0 & 5.0 & 4.0 & 1.0 & 3.0 & 2.0  \\
Coil100g30  & 5.0 & 6.0 & 4.0 & 2.0 & 3.0 & 1.0  \\
Coil100g40  & 6.0 & 5.0 & 4.0 & 2.0 & 3.0 & 1.0  \\
\midrule
Average rank~~&
~5.6667~ & ~5.3333~ & 4.0000 & 1.6667 & 3.0000 & \textbf{1.3333} \\
\bottomrule
\end{tabular}
}
\label{TableCoilRank}
\end{center}
\end{table}

\subsubsection{USPS data set}
\begin{figure}[htpb]
\begin{center}{{
\resizebox*{6.5cm}{!}
{\includegraphics{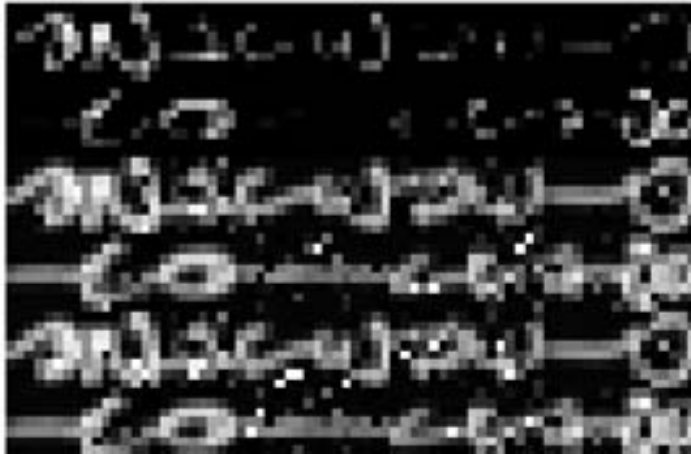}}}\hspace{5pt}
\caption{Original and noise sample images from USPS data set.}
\label{USPSsample}}
\end{center}
\end{figure}

\begin{figure*}[htpb]
\begin{center}{
\subfigure[Original data]{
\resizebox*{5.7cm}{!}
{\includegraphics{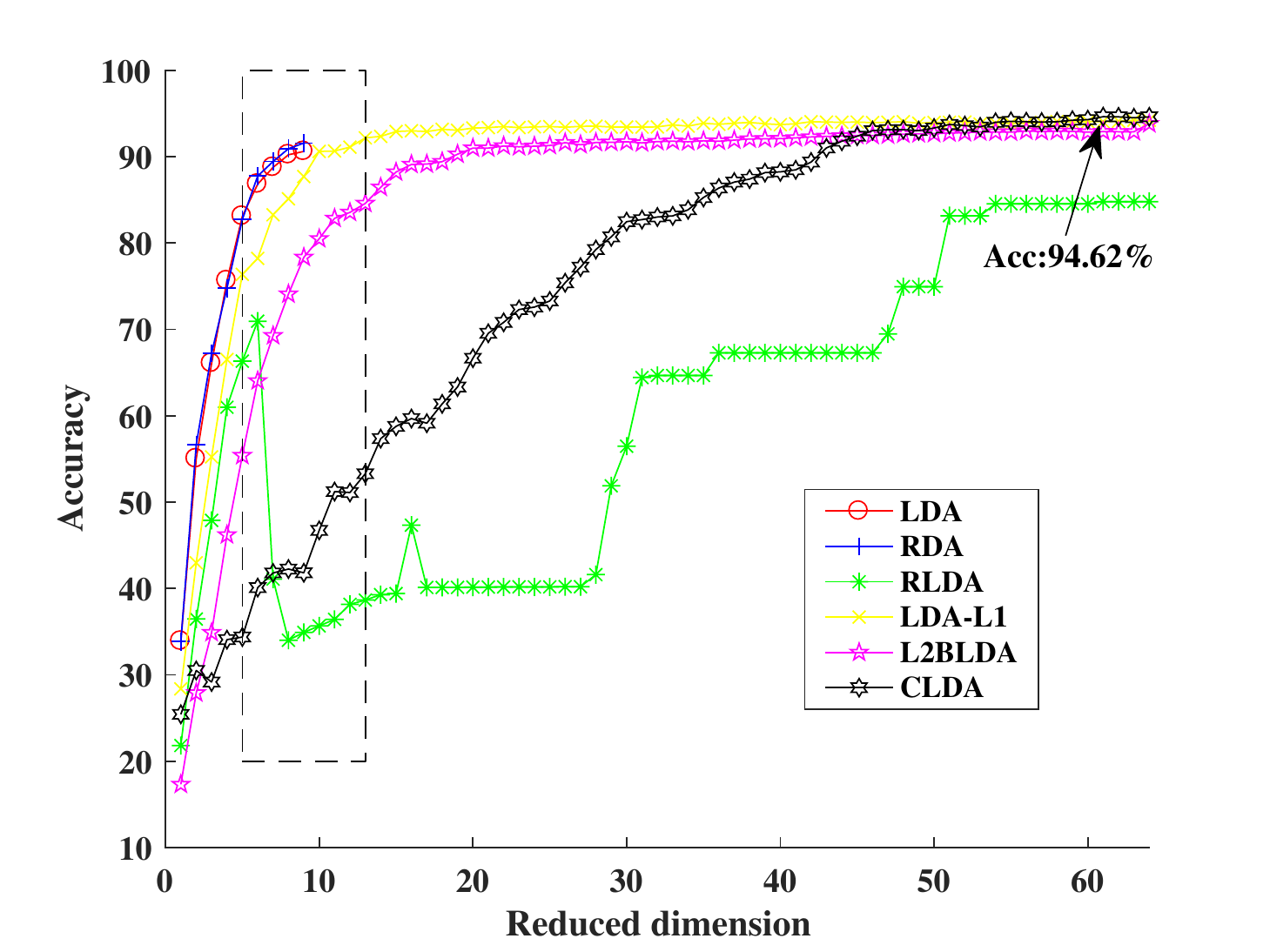}}}\hspace{5pt}
\subfigure[30\% noise data]{
\resizebox*{5.7cm}{!}
{\includegraphics{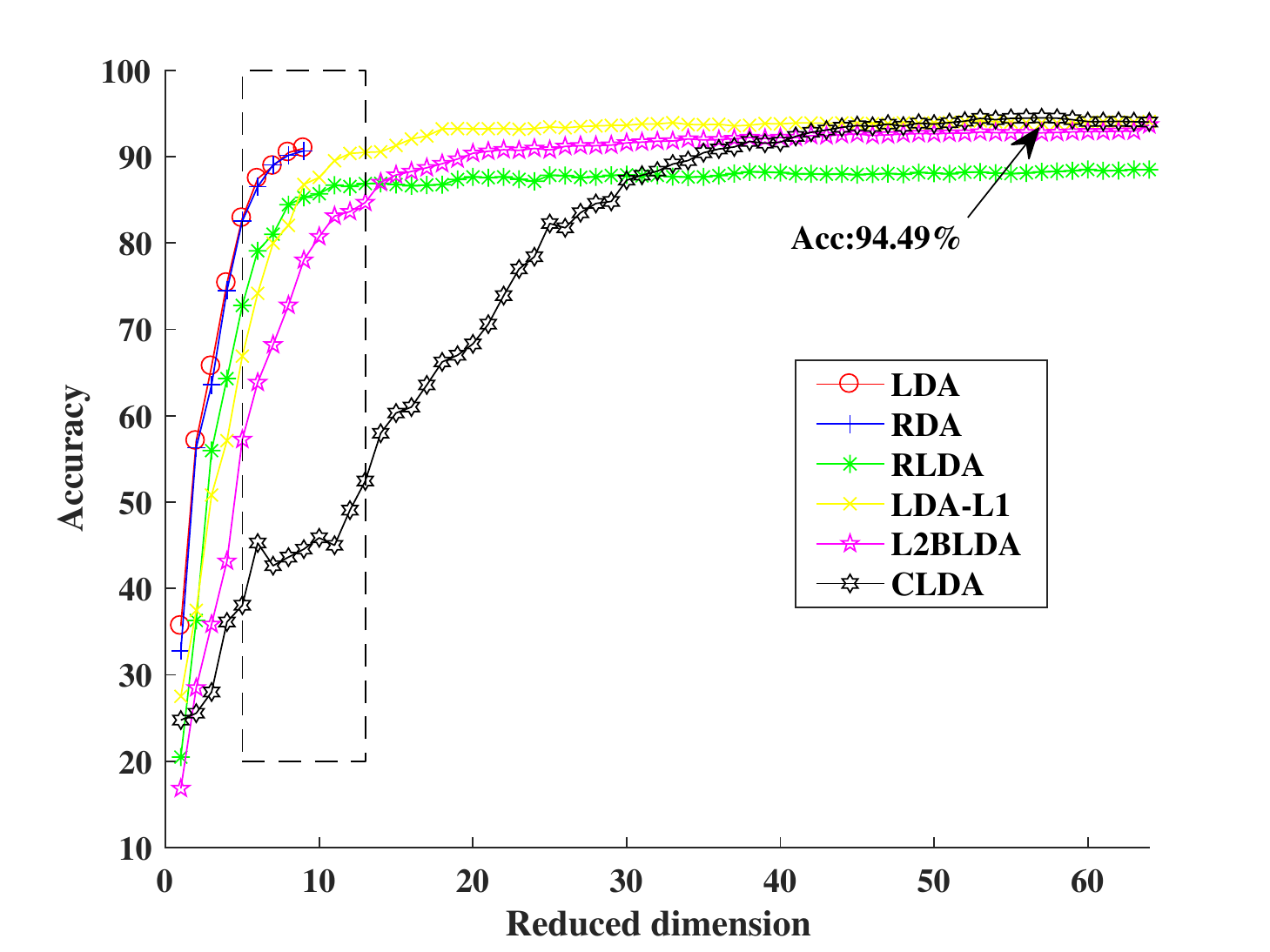}}}\hspace{5pt}
\subfigure[40\% noise data]{
\resizebox*{5.7cm}{!}
{\includegraphics{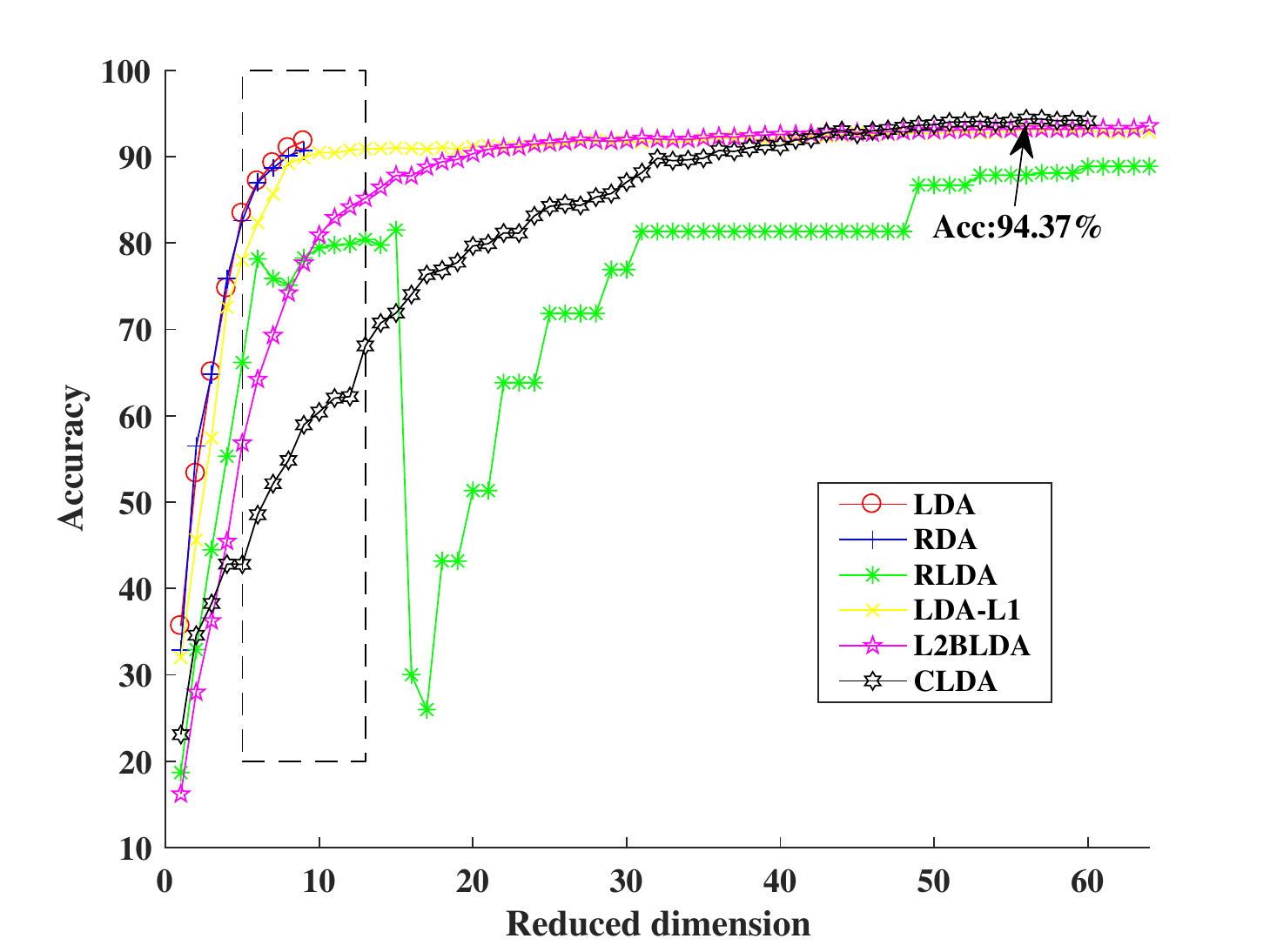}}}\hspace{5pt}
\caption{The variation of accuracies along different dimensions on USPS database.}
\label{USPSdim}}
\end{center}
\end{figure*}

We then consider USPS handwritten data set. USPS data set has 9298 digits images containing numbers 0-9, with each of them constituting a class. Each image is resized to 8$\times$8 pixel. Random 50\% samples from each class are used for training, while the rest data are used for testing.
To investigate the robustness of the proposed method, we further add
`salt \& pepper' of noise density 0.05 on each training sample that covers 30\% and 40\% rectangular area of the image, as shown in Figure \ref{USPSsample}.
We apply each dimensionality method on the above original and polluted training data, and obtain projection matrix. The classification results on test data are shown in Table \ref{TableUSPS}, and the corresponding rank results are listed in Table \ref{TableUSPSRank}. From the tables, we see our CLDA outperforms other methods.
To see the influence of reduced dimension to accuracy, we depict the variation of accuracies along dimensions in Fig.\ref{USPSdim}. The highest accuracy on each data is also shown in the figure. From the figure, we see as the number of reduced dimensions increases, the accuracies of all methods have general upward trend. Also, selecting an optimal dimension is important to all methods. For all three cases, the proposed CLDA owns the highest accuracies under its optimal dimensions.

\begin{table}[htbp]
\begin{center}
\caption{Classification results on USPS data.}
\resizebox{3.3in}{!}
{
\begin{tabular}{l|ccccccccccc}
\toprule
Data set & LDA ~&RDA~&RLDA~&LDA-L1~&L2BLDA~&CLDA\\
\midrule
USPSori~  & ~90.60~ & 91.56 & 84.79 & 94.22 & 93.81 & \textbf{94.62} \\
USPSs30~  & ~90.97~ & 90.66 & 88.52 & 94.13 & 93.68 & \textbf{94.49} \\
USPSs40~  & ~91.78~ & 90.69 & 88.88 & 93.12 & 93.59 & \textbf{94.37} \\
\bottomrule
\end{tabular}
}
\label{TableUSPS}
\end{center}
\end{table}

\begin{table}[htbp]
\begin{center}
\caption{Classification ranks on USPS data.}
\resizebox{3.3in}{!}
{
\begin{tabular}{l|ccccccccccc}
\toprule
{\multirow{2}{*}{Data set}} & LDA& RDA &RLDA~~&LDA-L1~~&L2BLDA&CLDA\\
& Rank& Rank& Rank& Rank& Rank& Rank\\
\midrule
USPSori  & 5.0 & 4.0 & 6.0 & 2.0 & 3.0 & 1.0  \\
USPSs30  & 4.0 & 5.0 & 6.0 & 2.0 & 3.0 & 1.0  \\
USPSs40  & 4.0 & 5.0 & 6.0 & 3.0 & 2.0 & 1.0  \\
\midrule
Average rank~~&
~4.3333~ & ~4.6667~ & 6.0000 & 2.3333 & 2.6667 &\textbf{ 1.0000} \\
\bottomrule
\end{tabular}
}
\label{TableUSPSRank}
\end{center}
\end{table}

\section{Conclusion}\label{secCon}

This paper introduced a capped $l_{p,q}$-norm for a matrix, where $p,q>0$, and proposed a novel linear discriminant analysis based on $l_{2,1}$-norm, named CLDA. Capped $l_{2,1}$-norm brought robustness to CLDA, and CLDA can be viewed as a weighted LDA. For a given $\epsilon$, the objective of CLDA was proved convergent.
Experimental results showed that compared to related LDAs, CLDA can effectively remove extreme outliers and suppress the effect of noise data. How to determine an appropriate thresholding parameter in CLDA and investigate a more efficient algorithm are our future works. Extending CLDA to matrix and tensor data is also interesting. The corresponding Matlab code for CLDA can be downloaded from http://www.optimal-group.org/Resources/Code/CLDA.html.

\section*{Acknowledgment}
This work is supported by the National Natural Science Foundation of China (No. 62066012, No.61703370, No.11871183, and No.61866010).

%





\end{document}